\documentclass[letterpaper]{article} 
\usepackage{aaai2027} 
\nocopyright
\usepackage[hyphens]{url}  
\usepackage{graphicx} 
\def\UrlFont{\rm}  
\usepackage{natbib}  
\usepackage{caption} 
\usepackage{algorithm}
\usepackage{algorithmic}
\usepackage{xcolor}

\usepackage{newfloat}
\usepackage{listings}
\DeclareCaptionStyle{ruled}{labelfont=normalfont,labelsep=colon,strut=off} 
\floatstyle{ruled}
\newfloat{listing}{tb}{lst}{}
\floatname{listing}{Listing}

\usepackage{booktabs}

\usepackage{multirow}
\usepackage{amsfonts}
\usepackage{amsmath}
\usepackage{xcolor}

\title{QuerySplat: Decoupling Geometry and Appearance Representations \\in 3DGS Prediction}

\author{
    Yinglong Li\textsuperscript{\rm 1,2}\equalcontrib,
    Donghui Shen\textsuperscript{\rm 2}\equalcontrib,
    Xiaoyu Zhang\textsuperscript{\rm 2},
    Zhichao Ye\textsuperscript{\rm 2},\\
    Hongyu Wu\textsuperscript{\rm 1}\corresponding,
    Aimin Hao\textsuperscript{\rm 1},
    Guofeng Zhang\textsuperscript{\rm 2,3}\corresponding,
    Haomin Liu\textsuperscript{\rm 2}
}
\affiliations{
    \textsuperscript{\rm 1}
    State Key Laboratory of Virtual Reality Technology and Systems,
    Beihang University\\
    \textsuperscript{\rm 2}
    InSpatio Research \quad
    \textsuperscript{\rm 3}
    State Key Lab of CAD\&CG,
    Zhejiang University
}

\begin{document}

\maketitle

\begin{abstract}
While feed-forward 3D Gaussian Splatting (3DGS) enables efficient 3D reconstruction, achieving high-fidelity rendering remains challenging. Existing pixel-aligned approaches suffer from spatial inflexibility and massive structural redundancy, whereas query-based methods lack 3D priors and entangle geometry with appearance, yielding blurry, pose-dependent results. To overcome these deficiencies, we propose \textbf{QuerySplat}, a feed-forward 3DGS framework driven by geometric priors and explicit appearance decoupling. Specifically, we design a dual-branch query-based decoder: the geometry branch leverages a pretrained Vision Geometric Model for spatial understanding, which intrinsically endows QuerySplat with pose-free modeling capabilities, while the appearance branch recovers high-frequency details through a dedicated pathway separated from geometric attribute regression. Extensive experiments demonstrate that QuerySplat mitigates the blurry rendering issues of early query-based models and consistently outperforms pixel-aligned approaches in rendering fidelity. On the challenging DL3DV benchmark, it achieves state-of-the-art novel view synthesis performance, with average PSNR gains of 2.30 dB and 1.04 dB over the best pose-free and pose-required baselines, respectively. Project Page: \textcolor{blue}{\url{https://inspatio.github.io/querysplat}}.

\end{abstract}


\begin{figure*}[t]
    \centering
    \includegraphics[width=0.88\textwidth]{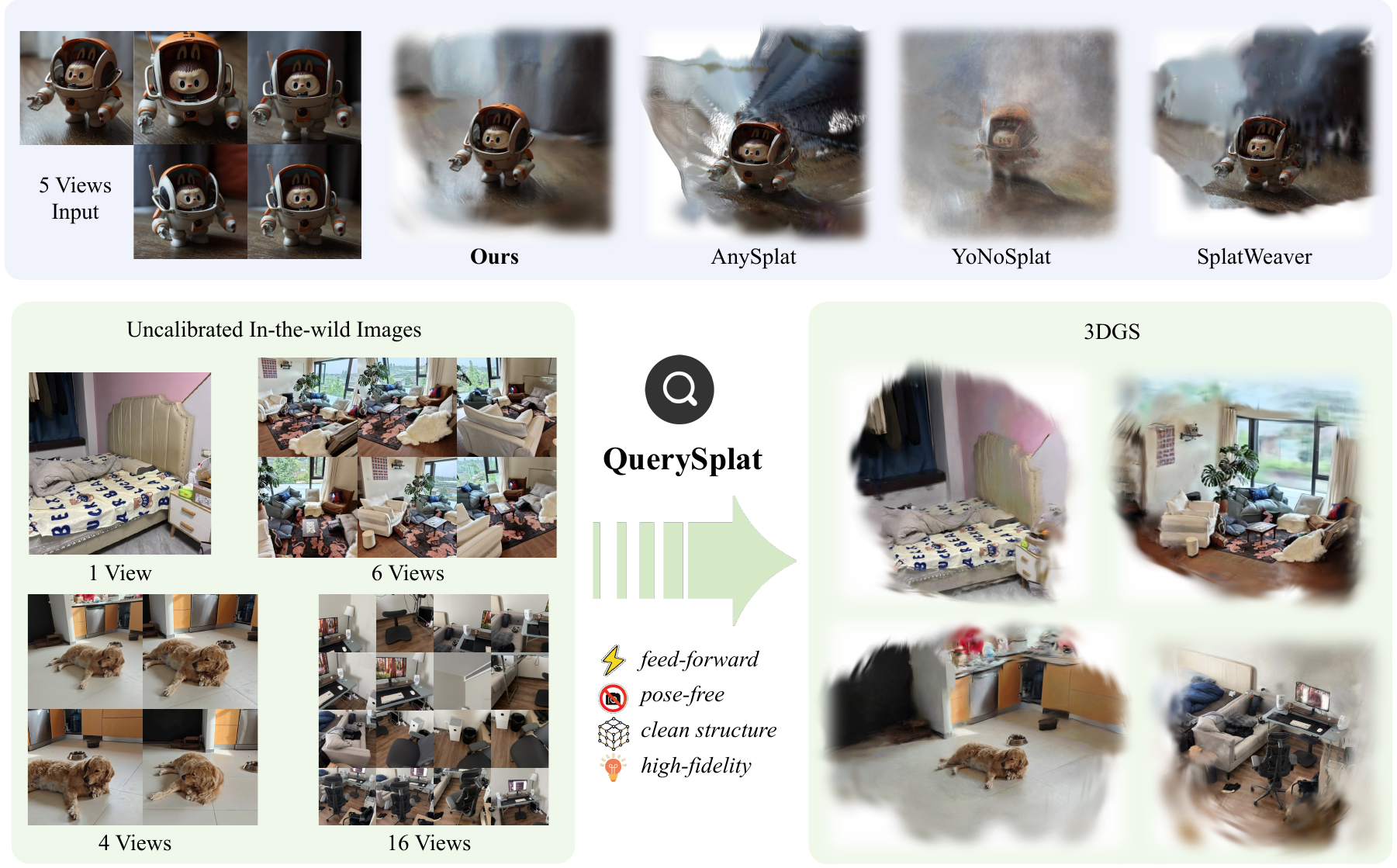}
    \caption{\textbf{QuerySplat overview.} Given uncalibrated images with varying view counts, QuerySplat reconstructs clean and high-fidelity 3D Gaussian scenes in a pose-free, feed-forward, and non-pixel-aligned manner. Compared with prior methods, it produces substantially more coherent scene structure while supporting in-the-wild inputs within seconds.}
    \label{fig:teaser}
\end{figure*}

\section{Introduction}
Recently, Feed-Forward Reconstruction (FFR) of 3D Gaussian Splatting (3DGS) \cite{jiang2025anysplat,ye2025yonosplat,ren2026tokengs,mescheder2025sharp} has emerged as a prominent research direction. By predicting parameters in a single pass, FFR preserves rapid inference while circumventing tedious per-scene optimization and dense multi-view captures, opening new possibilities for flexible 3D content creation.

Mainstream feed-forward methods primarily adopt a pixel-aligned generation paradigm, binding Gaussian primitives to camera rays. Although this design offers a simple image-to-Gaussian interface, strictly binding Gaussian primitives to the 2D observation space severely restricts their spatial degrees of freedom. Consequently, they are highly susceptible to depth errors and camera pose perturbations. Furthermore, multi-view pixel conflicts frequently accumulate into geometric shifts and ghosting artifacts, limiting the capacity to learn stable 3D structures. 

To overcome these constraints, query-based methods like TokenGS~\cite{ren2026tokengs} utilize learnable queries to decode Gaussians in continuous space. This spatial decoupling unbinds the primitive count from input resolutions, eliminates structural redundancy, and enables flexible allocation to complex regions. However, two critical limitations persist. First, existing methods entangle all Gaussian attributes within a unified query representation, ignoring their distinct modeling requirements: geometric attributes rely on global spatial reasoning, whereas appearance attributes depend on local high-frequency textures. Forcing them together creates a fundamental mismatch, where appearance variations disrupt geometric learning, causing over-smoothed renderings. Second, relying solely on 2D photometric supervision to learn structures from scratch leaves these models without explicit 3D priors. Lacking a canonical coordinate system, they remain rigidly dependent on exact camera poses, hindering unposed, in-the-wild applicability.

Based on these observations, we propose QuerySplat, an attribute-decoupled, high-fidelity FFR 3DGS framework. At its core is an attribute-aware dual-query decoder. The geometry branch leverages dedicated queries to predict spatial attributes, while the appearance branch focuses solely on high-frequency details. This dual-branch decoupling in both feature and query spaces allows each attribute to independently aggregate essential information, achieving well-defined geometry alongside rich textures.

To provide robust spatial support, we introduce a pretrained Vision Geometric Model (VGM) as a universal prior. It injects stable 3D structural features and establishes a canonical coordinate system, fundamentally enabling pose-free reconstruction. Additionally, we leverage VGM-predicted depth to formulate a transient early-stage regularization strategy. Combined with an opacity-floor constraint, this regularization stabilizes Gaussian initialization while preserving the queries' flexibility to freely reorganize under later image-space supervision.


Extensive experiments demonstrate that QuerySplat achieves state-of-the-art novel view synthesis performance on the large-scale and challenging DL3DV benchmark, consistently outperforming existing methods. The generated scenes exhibit superior geometric coherence, sharper object boundaries, and richer texture details, validating the effectiveness of the attribute-decoupled query mechanism.

The main contributions are summarized as follows:


\begin{itemize}
    \item We propose \textbf{QuerySplat}, an attribute-aware dual-query framework. By designing dedicated query and feature pathways for different Gaussian attributes, it enables the harmonious joint modeling of well-defined 3D structures and high-frequency appearances.

    \item We introduce a VGM-guided geometry branch that leverages early-stage regularization to stabilize non-pixel-aligned 3DGS geometry prediction, reducing learning uncertainty and enabling pose-free scene reconstruction.

    \item Extensive experiments demonstrate that QuerySplat achieves consistent improvements in both visual fidelity and geometric organization, validating the effectiveness of the attribute-decoupled query paradigm for addressing the challenges of non-pixel-aligned feed-forward 3DGS.
\end{itemize}

\section{Related Work}

\subsection{3D Gaussian Splatting}
3D Gaussian Splatting (3DGS)~\cite{kerbl20233dgs} offers high-fidelity real-time rendering but requires time-consuming per-scene optimization~\cite{yu2024gof, chen2024pgsr}. To address this, Feed-Forward Reconstruction (FFR) methods~\cite{szymanowicz2024splatterimage, charatan2024pixelsplat, chen2024mvsplat, li2026tokensplat, xu2024freesplatter, gupta2026GenWildSplat, moreau2025offthegrid, wang2026trisplat} have been proposed to infer scenes directly from sparse views in a single pass. Recent advancements further improve generalizability using explicit geometric priors~\cite{xu2025depthsplat, jiang2025anysplat}, pose-free formulations~\cite{ye2025yonosplat, huang2025spfsplat}, or adaptive primitive allocation~\cite{wan2026splatweaver}. Despite these efficiency gains, mainstream FFR models strictly anchor Gaussian primitives to camera rays. This pixel-aligned paradigm fundamentally restricts spatial degrees of freedom, rendering models vulnerable to noisy camera poses and limiting their capacity to represent complex geometries.


\subsection{Query-based 3D Scene Representation}
To overcome pixel-aligned constraints, query-based methods utilize learnable tokens to aggregate spatial information. Originating in 2D vision~\cite{carion2020detr}, this paradigm has successfully extended to 3D perception~\cite{wang2022detr3d, schult2023mask3d} and scene reconstruction via voxel~\cite{li2023voxformer} or triplane~\cite{hong2024lrm} representations. Recently, TokenGS~\cite{ren2026tokengs} introduced this mechanism to FFR 3DGS, completely breaking free from 2D pixel grids. However, while releasing spatial degrees of freedom, learning both complex geometry and fine-grained textures jointly from unified queries causes severe attribute entanglement. This optimization bottleneck degrades rendering quality—yielding over-smoothed, blurry results—and maintains a rigid dependency on exact camera poses.


\subsection{Vision Geometric Models}
Vision Geometric Models (VGMs) pretrained on massive datasets have demonstrated strong multi-view spatial reasoning capabilities. Pioneered by DUSt3R~\cite{wang2024dust3r} and MASt3R~\cite{leroy2024mast3r} for pose-free point-map regression, this paradigm recently advanced through highly performant architectures like VGGT~\cite{wang2025vggt}, Pi3~\cite{wang2025pi3}, DA3~\cite{lin2025da3}, and VGGT-$\Omega$~\cite{wang2026vggtomega}. To scale these foundational models to large environments, methods like ZipMap~\cite{jin2026zipmap} and Scal3R~\cite{xie2026scal3r} integrate test-time training (TTT) for global consistency, whereas LingBot-Map~\cite{chen2026lingbot-map} pursues a purely feed-forward streaming paradigm without post-optimization overhead. Despite these rapid architectural and system-level advancements, the integration of VGMs into downstream 3D generation pipelines remains relatively shallow.

\begin{figure*}[t]
    \centering
    \includegraphics[width=0.95\textwidth]{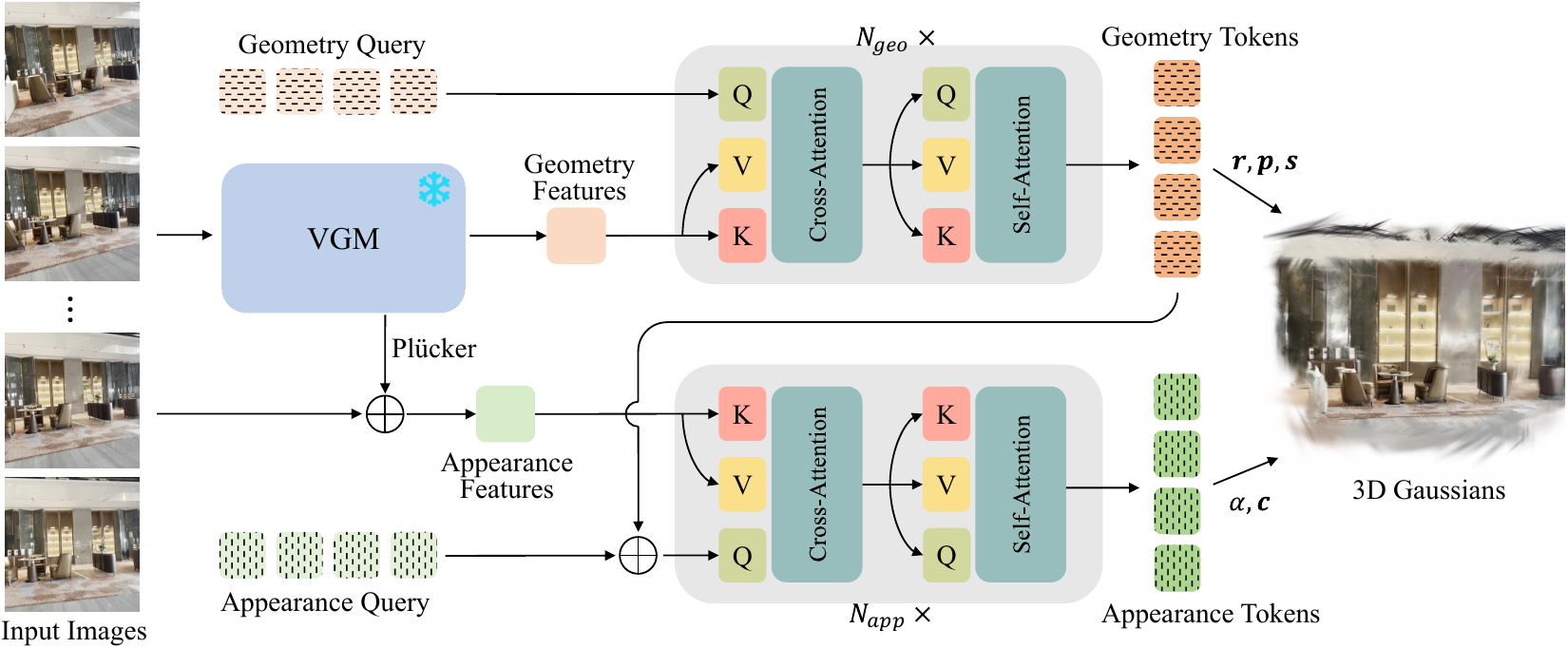}
    \caption{\textbf{Method Overview.} A Vision Geometric Model (VGM) encodes geometry-aware memory and defines cameras. Geometry queries decode spatial Gaussian parameters from VGM features, while appearance queries read RGB/Pl\"ucker memory to predict the others. The resulting Gaussians are supervised by differentiable splatting in the VGM-defined coordinate system.}
    \label{fig:method_overview}
\end{figure*}

\section{Method}
\label{sec:method}


Given a set of input images, our goal is to predict a renderable 3D Gaussian representation in a single feed-forward pass. The core of QuerySplat is an attribute-aware dual-branch, dual-query decoder that explicitly decouples geometry and appearance modeling. Building upon this design, we introduce a pretrained Vision Geometric Model (VGM) to provide geometry-aware features, camera estimates, and a consistent coordinate system. 

Formally, given input views $\mathcal{I}_{\mathrm{in}}=\{I_i\}_{i=1}^{N}$, the network predicts a Gaussian set
\begin{equation}
    \mathcal{G}
    =
    F_{\theta}(\mathcal{I}_{\mathrm{in}})
    =
    \{g_k\}_{k=1}^{K},
    \ \ 
    g_k
    =
    (\mathbf{r}_k,\mathbf{p}_k,\mathbf{s}_k,\alpha_k,\mathbf{c}_k),
\end{equation}
where $\mathbf{r}$, $\mathbf{p}$, $\mathbf{s}$, $\alpha$, and $\mathbf{c}$ denote the rotation quaternion, center, anisotropic scale, opacity, and spherical-harmonic coefficients, respectively. Instead of producing pixel-aligned primitives, the model employs learnable Gaussian queries as scene-level slots. Each query is decoded into a group of Gaussian primitives, which are supervised through differentiable rendering under VGM-predicted cameras. An overview of the framework is provided in Figure~\ref{fig:method_overview}.

\subsection{Pretrained VGM Features}
\label{sec:frozen_vgm}

A VGM is a large-scale pretrained multi-view model that can infer geometry-aware features, camera poses, intrinsics, and optionally dense depth. For the concrete VGGT-$\Omega$~\cite{wang2026vggtomega} instantiation, each input view is represented by a camera token, register/scene tokens, and dense patch tokens. We use them to construct geometric features $\mathbf{F}_{\mathrm{geo}}$, where patch tokens are extracted from four intermediate layers and fused with a lightweight learnable layer mixer~\cite{cao2026vggtdet}, while the others are taken only from the final layer.

The VGM is frozen throughout training. This is important because large-scale VGMs already encode strong multi-view geometry priors; jointly finetuning them with the Gaussian generator may damage their camera and geometry consistency. Freezing the VGM preserves this prior and cleanly separates the roles of the two parts: the VGM performs generic geometric reasoning, while the trainable decoder learns to translate $\mathbf{F}_{\mathrm{geo}}$ into the geometry of Gaussian primitives.

\subsection{Self-Calibrated Coordinate System}
\label{sec:vgm_native_coords}

By deriving cameras and scene coordinates from the VGM, our pipeline reduces reliance on external camera annotations and supports training on heterogeneous or RGB-only image collections. We first process only the input views to extract $\mathbf{F}_{\mathrm{geo}}$ and define the coordinate frame for geometry decoding. For rendering supervision, we separately process the union of input and supervision views to estimate their cameras, without sharing features between the two passes. Crucially, we maintain strict isolation between these two passes with no feature exchange, explicitly preventing any target-view information leakage into the reconstruction pipeline. Since the resulting frames may differ by a similarity gauge, we align the all-view cameras to the input-only frame using a $\mathrm{Sim}(3)$ transformation estimated from the shared input views. This places $\mathbf{F}_{\mathrm{geo}}$, Pl\"ucker rays, Gaussian centers, and supervision cameras in a common coordinate system.

\subsection{Decoupled Queries}
\label{sec:decoupled_queries}

Our decoder separates the question of \emph{where} Gaussians should be placed from the question of \emph{what} they should look like. The motivation is simple. Geometry Features are strong at camera-consistent layout and scene-level structure, but they are not optimized to preserve all high-frequency information. Appearance features retain local texture and ray information, but injecting them too early into the geometry stream can let appearance gradients disturb stable Gaussian placement. Thus, we keep geometry decoding spatially driven and defer appearance modeling to a separate branch.

\paragraph{Geometry Queries.}
A set of learnable Geometry Queries $\mathbf{Q}_{\mathrm{geo}}$ attends to $\mathbf{F}_{\mathrm{geo}}$ through transformer decoder blocks. Each block contains cross-attention to Geometry Features, self-attention among queries, and an MLP update. The decoded tokens are
\begin{equation}
    \mathbf{Z}_{\mathrm{geo}}=D_{\mathrm{geo}}(\mathbf{Q}_{\mathrm{geo}},\mathbf{F}_{\mathrm{geo}}),
\end{equation}
which are mapped by a geometry head to Gaussian centers, scales, and rotations. Because the queries are not tied to pixels, each query behaves like a latent scene slot that can collect evidence from multiple views and feature locations before being expanded into Gaussian primitives.

\paragraph{Appearance Queries.}
For appearance, we construct Appearance Features $\mathbf{F}_{\mathrm{app}}$ from RGB patch embeddings and Pl\"ucker ray embeddings. The rays are computed using the input cameras predicted by the input-only VGM pass, so the appearance stream remains in the same VGM-native frame as the geometry stream. The appearance decoder takes the geometry tokens $\mathbf{Z}_{\mathrm{geo}}$ as its base query state, augments them with learnable Appearance Queries $\mathbf{Q}_{\mathrm{app}}$, and attends to $\mathbf{F}_{\mathrm{app}}$. The decoded tokens are
\begin{equation}
    \mathbf{Z}_{\mathrm{app}}
    =
    D_{\mathrm{app}}(\mathbf{Z}_{\mathrm{geo}},\mathbf{Q}_{\mathrm{app}},\mathbf{F}_{\mathrm{app}}),
\end{equation}
which are mapped to Gaussian opacities and SH colors. 
Finally, we assemble both into the full Gaussian set $\mathcal{G}$.

\subsection{Loss Functions}
\label{sec:objectives}

The predicted Gaussians are rendered by differentiable Gaussian splatting under the aligned supervision cameras. We train the model with
\begin{equation}
\label{eq:full_loss}
    \mathcal{L}
    =
    \mathcal{L}_{\mathrm{photo}}
    +
    \lambda_{\mathrm{vis}}\mathcal{L}_{\mathrm{vis}}
    +
    \beta_{\mathrm{cd}}(t)\mathcal{L}_{\mathrm{cd}}
    +
    \beta_{\alpha}(t)\mathcal{L}_{\alpha},
\end{equation}
where $\beta_{\mathrm{cd}}(t)$ and $\beta_{\alpha}(t)$ are stage-dependent schedules used only for early training regularization.

\noindent\textbf{Photometric reconstruction.}
This is the main rendering signal. We combine pixel-level, structural, and perceptual reconstruction terms:
\begin{equation}
    \mathcal{L}_{\mathrm{photo}}
    =
    \mathcal{L}_{1}
    +
    \lambda_{\mathrm{ssim}}\mathcal{L}_{\mathrm{ssim}}
    +
    \lambda_{\mathrm{lpips}}\mathcal{L}_{\mathrm{lpips}} .
\end{equation}
The L1 term enforces pixel accuracy, SSIM stabilizes local structure, and LPIPS improves perceptual texture quality.

\noindent\textbf{Visibility regularization.}
Following TokenGS~\cite{ren2026tokengs}, we penalize Gaussian centers that fall outside all relevant camera frusta or behind cameras, preventing latent queries from producing floating Gaussians that receive little rendering gradient. In our setting, we evaluate it on the relevant VGM-predicted cameras, including input and supervision views when available.

\noindent\textbf{Early-stage regularization.}
At the beginning of base training, we introduce two temporary regularizers to stabilize Gaussian initialization. First, the depth predicted by the input-only VGM pass is back-projected into a pseudo point cloud $\mathcal{P}$, and the predicted Gaussian centers $\mathcal{G}_{p}$ are regularized by a bidirectional Chamfer distance:
\begin{equation}
    \mathcal{L}_{\mathrm{cd}}
    =
    d_{\mathrm{CD}}(\mathcal{G}_{p},\mathcal{P}).
\end{equation}
Second, we use an opacity-floor regularizer to prevent Gaussians from becoming transparent too early:
\begin{equation}
    \mathcal{L}_{\alpha}
    =
    \mathbb{E}_{\alpha\in\mathcal{A}}
    \left[
    \max\left(
    0,
    \log \alpha_{\min}
    -
    \log(\max(\alpha,\epsilon))
    \right)
    \right],
\end{equation}
where $\mathcal{A}$ denotes the predicted Gaussian opacities, $\alpha_{\min}$ is the minimum opacity floor, and $\epsilon$ is a small constant for numerical stability. Both terms serve only as transient initialization priors and are gradually removed. We find that enforcing them throughout training destabilizes optimization, since the rendering-optimal Gaussian configuration generally departs from the depth-derived point cloud and requires both positions and opacities to be freely reorganized under image-space supervision.

\begin{table*}[!t]
  \centering
  \fontsize{9pt}{10pt}\selectfont
  \setlength{\tabcolsep}{2.5pt}
  \begin{tabular*}{\textwidth}{@{\extracolsep{\fill}}lc*{9}{c}@{}}
    \toprule
    \multirow{2}{*}{Method} & \multirow{2}{*}{Pose-free} & \multicolumn{3}{c}{2 views} & \multicolumn{3}{c}{4 views} & \multicolumn{3}{c}{12 views} \\
    \cmidrule(lr){3-5}\cmidrule(lr){6-8}\cmidrule(lr){9-11}
    & & PSNR $\uparrow$ & SSIM $\uparrow$ & LPIPS $\downarrow$ & PSNR $\uparrow$ & SSIM $\uparrow$ & LPIPS $\downarrow$ & PSNR $\uparrow$ & SSIM $\uparrow$ & LPIPS $\downarrow$ \\
    \midrule
    DepthSplat~\shortcite{xu2025depthsplat} & $\times$ & \underline{20.5357} & 0.6721 & \underline{0.2592} & 22.9079 & \underline{0.7683} & \underline{0.1853} & 21.6785 & 0.7549 & \underline{0.2050} \\
    TokenGS~\shortcite{ren2026tokengs} & $\times$ & 20.4258 & \underline{0.6736} & 0.3783 & \underline{23.2704} & 0.7565 & 0.3026 & 21.8094 & 0.6960 & 0.3696 \\
    YoNoSplat~\shortcite{ye2025yonosplat} & $\times$ & 19.8356 & 0.6296 & 0.2909 & 22.8488 & 0.7469 & 0.1987 & \underline{22.7326} & \underline{0.7608} & \textbf{0.1958} \\
    \midrule
    AnySplat~\shortcite{jiang2025anysplat} & $\checkmark$ & 14.0742 & 0.4447 & 0.4475 & 17.3938 & 0.5483 & 0.3418 & 19.6253 & 0.6334 & 0.2989 \\
    NoPoSplat~\shortcite{ye2025noposplat} & $\checkmark$ & 19.0650 & 0.5997 & 0.3164 & -- & -- & -- & -- & -- & -- \\
    SPFSplat~\shortcite{huang2025spfsplat} & $\checkmark$ & 19.1163 & 0.5874 & 0.3046 & -- & -- & -- & -- & -- & -- \\
    SplatWeaver~\shortcite{wan2026splatweaver} & $\checkmark$ & 15.6953 & 0.4991 & 0.3666 & 19.2413 & 0.6294 & 0.2715 & 20.4047 & 0.6683 & 0.2515 \\
    YoNoSplat~\shortcite{ye2025yonosplat} & $\checkmark$ & 19.1675 & 0.5908 & 0.3081 & 21.9644 & 0.6983 & 0.2167 & 21.6323 & 0.6953 & 0.2184 \\
    \textbf{Ours} & $\checkmark$ & \textbf{21.3888} & \textbf{0.6990} & \textbf{0.2585} & \textbf{24.5765} & \textbf{0.8002} & \textbf{0.1843} & \textbf{23.7005} & \textbf{0.7685} & 0.2297 \\
    \midrule
    \textbf{Ours} + TTO-20 & $\checkmark$ & 22.0094 & 0.7168 & 0.2478 & 26.2524 & 0.8304 & 0.1591 & 27.0003 & 0.8472 & 0.1667 \\
    \textbf{Ours} + TTO-50 & $\checkmark$ & 21.9437 & 0.7149 & 0.2463 & 26.3779 & 0.8321 & 0.1537 & 27.7199 & 0.8615 & 0.1464 \\
    \bottomrule
  \end{tabular*}
  \caption{Interpolation results averaged over the large, medium, and small splits. Higher PSNR/SSIM and lower LPIPS are better. \textbf{Bold} and \underline{underlined} values indicate the best and second-best results, respectively; TTO variants are excluded from this comparison. See Appendix A for complete results for each view setting and split.}
  \label{tab:main-interpolation}
\end{table*}

\begin{figure*}[!t]
    \centering
    \includegraphics[width=\textwidth]{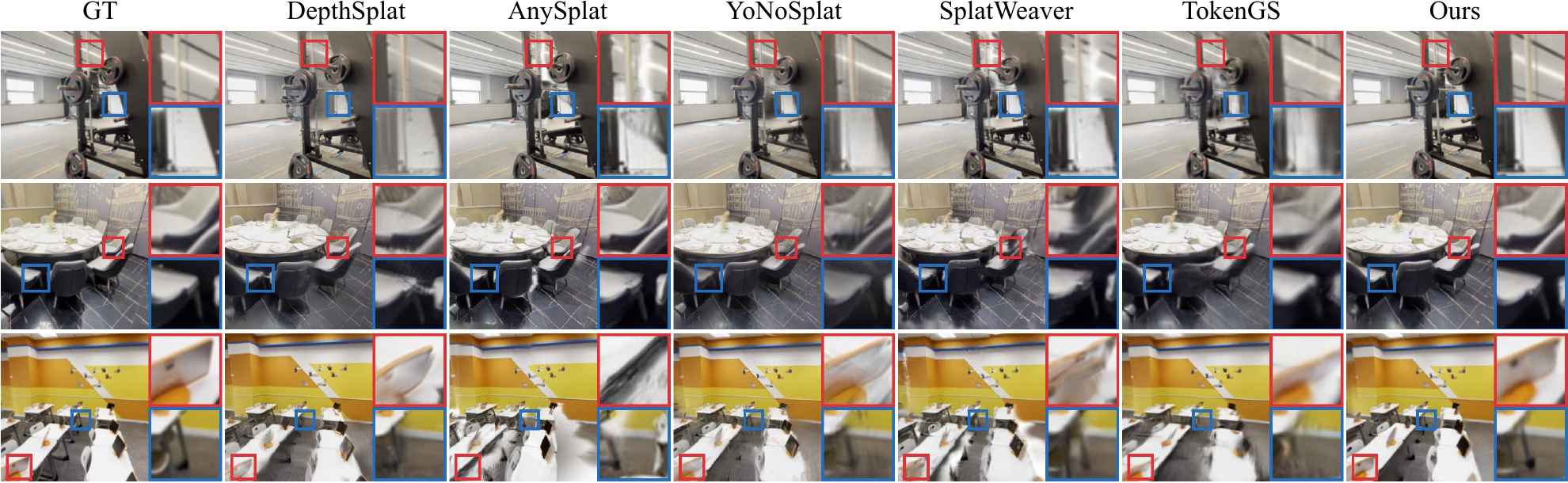}
    \caption{\textbf{Qualitative comparison.} QuerySplat preserves finer textures and sharper object boundaries, producing more faithful and visually detailed renderings than prior methods.}
    \label{fig:qualitative_comparison}
\end{figure*}

\begin{figure}[!t]
    \centering
    \includegraphics[width=\linewidth]{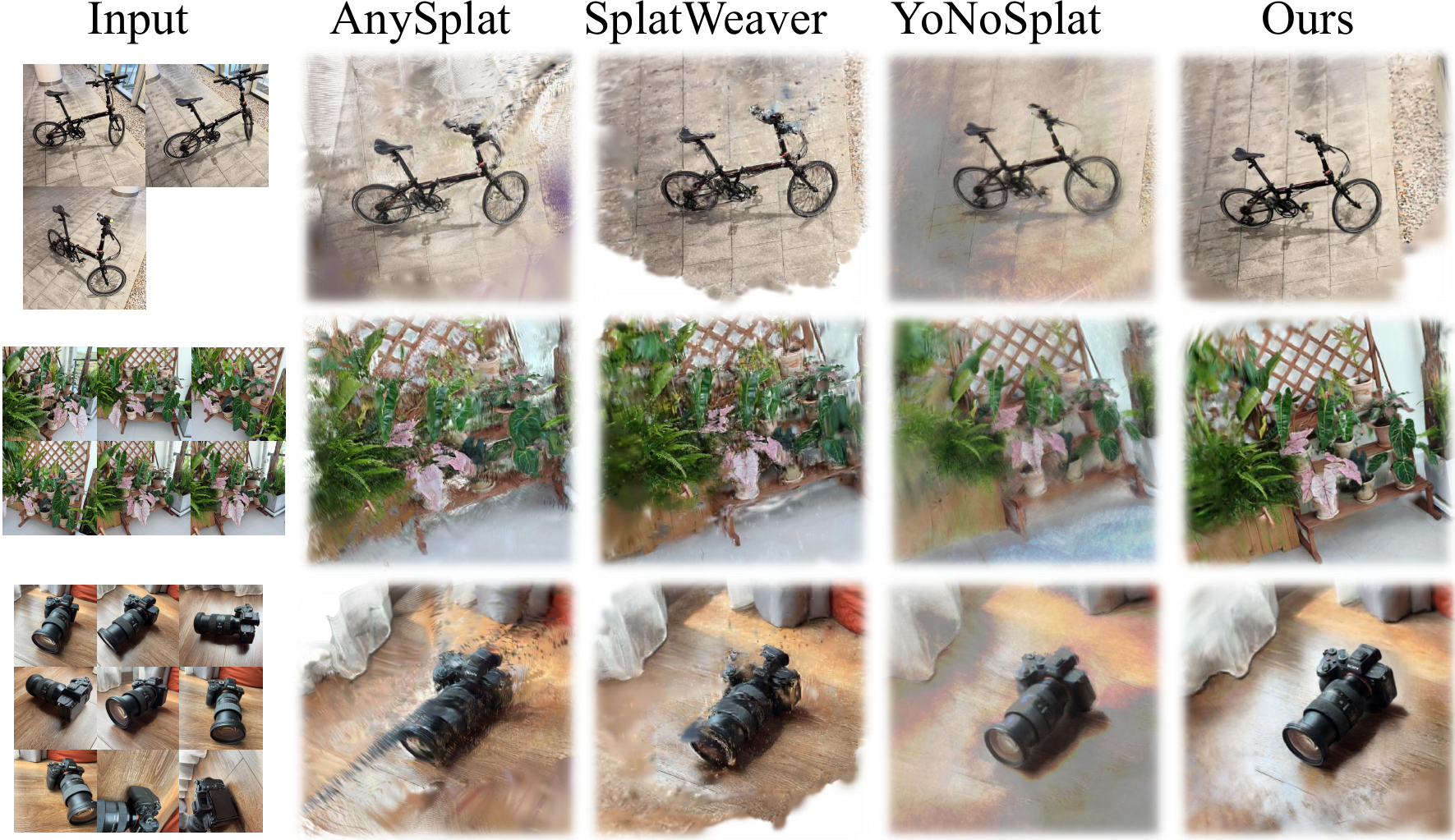}
    \caption{\textbf{In-the-wild qualitative comparison.} Each row compares 3DGS reconstructions from casually captured unposed images. QuerySplat yields cleaner geometry and sharper details than competing pose-free methods.}
    \label{fig:in_the_wild_comparison}
\end{figure}

\subsection{Optional Test-Time Optimization}
\label{sec:tto}

The main method is feed-forward, but it also supports lightweight test-time optimization (TTO). During TTO, we keep the learned queries, decoders, and output heads fixed, and optimize the extracted features with rendering losses on input views. This feature-space adaptation refines the scene representation while keeping it constrained by the pretrained query-to-Gaussian reconstruction model, providing a simple trade-off between runtime and reconstruction fidelity.

\section{Experiments}
\label{sec:experiments}

\subsection{Implementation Details}
\label{sec:imple_details}

We use pretrained VGGT-$\Omega$~\cite{wang2026vggtomega} as the VGM encoder, with $N_{\mathrm{geo}}=12$ and $N_{\mathrm{app}}=6$. Each query token predicts 64 Gaussian primitives. We train exclusively on DL3DV~\cite{ling2024dl3dv} using $512 \times 512$ center-cropped images. Training consists of a base and a progressive finetuning stage. Base training optimizes 1,024 queries across 4 random input views for 300K iterations with a learning rate of $10^{-4}$. The Chamfer distance and opacity regularizers are annealed to zero over the first 20K iterations. LPIPS is introduced after 100K iterations and gradually increased to $\lambda_{\mathrm{lpips}}=0.05$. During finetuning, we progressively double the number of queries to 8,192 (Newly added queries are initialized from pretrained ones with small perturbations $\mathbf{q}^{\mathrm{new}}_i=\mathbf{q}^{\mathrm{old}}_{i\bmod N_{\mathrm{old}}}+\boldsymbol{\xi}$), training each expansion for 30K iterations with a learning rate of $10^{-5}$ and randomly sampled 2--12 input views. Both stages use the AdamW~\cite{loshchilov2019adamw} optimizer, a global batch size of 64, and a 5\% linear warm-up followed by cosine decay. We set $\lambda_{\mathrm{ssim}}=0.2$ and $\lambda_{\mathrm{vis}}=1.0$. Training is conducted in BF16 on 64 NVIDIA A800 GPUs, with one sample per GPU. Complete training details and hyperparameter settings are provided in Appendix~E.

\subsection{Comparisons}
\label{sec:comparisons}

We compare QuerySplat with recent posed and pose-free feed-forward 3DGS methods, whose full list is provided in Table~\ref{tab:main-interpolation}.
For all baseline methods, we use the official implementations and released model weights. We construct evaluation cases from DL3DV-Evaluation~\cite{ling2024dl3dv} (independent of DL3DV-10K). For 2-, 4-, and 12-view settings, cases are divided into small, medium, and large splits according to the image interval, with 300 cases randomly sampled for each split. We evaluate on input, interpolation, and extrapolation views using PSNR, SSIM, and LPIPS. 
For fair comparison, inputs are resized to a shorter side of 256 pixels while preserving the aspect ratio. Following inference at each method's native resolution, the rendered outputs are resized back to the original resolution for metric computation. To strictly prevent target-view leakage, all pose-free baselines use the all-view pass solely for camera estimation and $\mathrm{Sim}(3)$ alignment, whereas posed methods receive input poses.
Table~\ref{tab:main-interpolation} reports interpolation results averaged over the three splits, with complete results provided in Appendix~A. We additionally report QuerySplat with 20 and 50 steps of test-time optimization to demonstrate the optional gains from feature-space adaptation; these variants are excluded from the comparison ranking. Without TTO, QuerySplat consistently outperforms existing methods and achieves the strongest overall performance across different input-view settings.


Beyond the quantitative results, Figure~\ref{fig:qualitative_comparison} shows that QuerySplat preserves fine textures and object boundaries, achieving visual sharpness comparable to pixel-aligned methods while avoiding the blurry outputs of previous query-based approaches. Figure~\ref{fig:in_the_wild_comparison} also demonstrates cleaner structures and sharper details than competing pose-free methods on casually captured inputs. More fundamentally, QuerySplat advances prior query-based reconstruction by preserving its clean and flexible scene representation while substantially improving both geometric organization and texture fidelity. Geometry-aware decoding produces more coherent Gaussian structures, and the dedicated appearance branch recovers fine details that earlier query-based methods tend to smooth out. As a result, QuerySplat closes the quality gap between query-based and pixel-aligned paradigms, achieving comparable visual sharpness without sacrificing the structural advantages of non-pixel-aligned reconstruction.

\subsection{Ablation Studies}
\label{sec:ablation}

Due to the prohibitive computational cost of fully training all model variants, ablation models are trained only in the base stage for 150K iterations and evaluated on 4-view interpolation averaged over the large, medium, and small splits.

\begin{figure}[!t]
    \centering
    \includegraphics[width=0.9\linewidth]{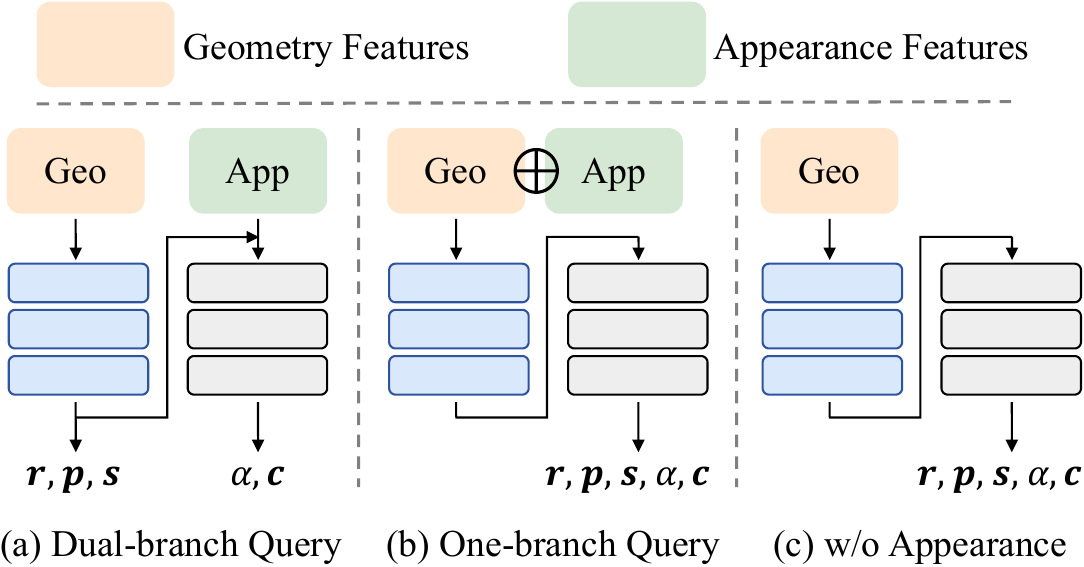}
    \caption{\textbf{Feature aggregation variants.} Illustration of (a) our dual-branch design, (b) one-branch feature fusion, and (c) geometry-only prediction without appearance features.}
    \label{fig:ablation_structure}
\end{figure}

\begin{table}[!t]
  \centering
  \fontsize{9pt}{10pt}\selectfont
  \setlength{\tabcolsep}{4pt}
  \begin{tabular}{l|ccc}
    \toprule
    Method & PSNR $\uparrow$ & SSIM $\uparrow$ & LPIPS $\downarrow$ \\
    \midrule
    \textbf{Ours} & \textbf{22.8963} & \textbf{0.7381} & \textbf{0.2682} \\
    \textbf{Ours} (One-branch Query) & \underline{20.9161} & \underline{0.6593} & \underline{0.3440} \\
    \textbf{Ours} (w/o Appearance) & 18.6447 & 0.5782 & 0.4254 \\
    \bottomrule
  \end{tabular}
  \caption{Feature aggregation ablation at 4-view interpolation, with early-stage regularization enabled.}
  \label{tab:ablation-appearance}
\end{table}

\begin{figure}[!t]
    \centering
    \includegraphics[width=\linewidth]{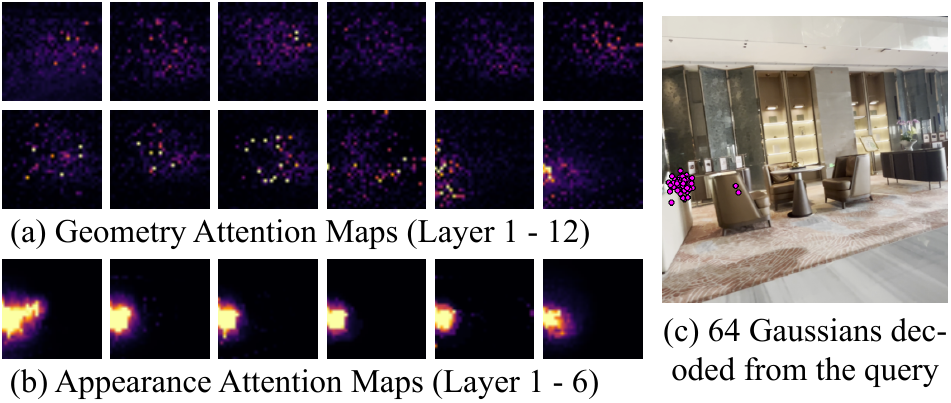}
    \caption{\textbf{Query attention visualization.} Panels (a) and (b) show geometry and appearance attention for one query, while (c) shows its decoded Gaussians projected onto the image.}
    \label{fig:attention_map}
\end{figure}

\paragraph{Attribute-aware feature aggregation.}
Figure~\ref{fig:ablation_structure} illustrates the three variants evaluated in Table~\ref{tab:ablation-appearance}, with matched layer and parameter counts. The results show that the benefit of our dual-branch design does not come from adding appearance features. Removing the appearance stream degrades reconstruction quality, while merging appearance features into the geometry stream performs worse. Instead, different Gaussian attributes should gather evidence suited to their prediction. Spatial parameters require broad cross-view context to establish a coherent scene layout, whereas the others rely on local texture and viewing-direction cues. The attention patterns in Figure~\ref{fig:attention_map} support this behavior: geometry queries attend to broader structural regions, while the appearance branch focuses on local evidence around the predicted Gaussians. Separating these aggregation processes preserves stable scene organization while recovering finer appearance details.

\begin{table}[!t]
  \centering
  \fontsize{9pt}{10pt}\selectfont
  \begin{tabular}{lc|ccc}
    \toprule
    Method & Early Reg. & PSNR $\uparrow$ & SSIM $\uparrow$ & LPIPS $\downarrow$ \\
    \midrule
    \textbf{Ours} & $\checkmark$ & \textbf{22.8963} & \textbf{0.7381} & \textbf{0.2682} \\
    \textbf{Ours} & $\times$ & 22.6090 & 0.7228 & 0.2877 \\
    \bottomrule
  \end{tabular}
  \caption{Ablation of early-stage regularization at 4-view interpolation.}
  \label{tab:ablation-regularization}
\end{table}

\begin{figure}[!t]
    \centering
    \includegraphics[width=\linewidth]{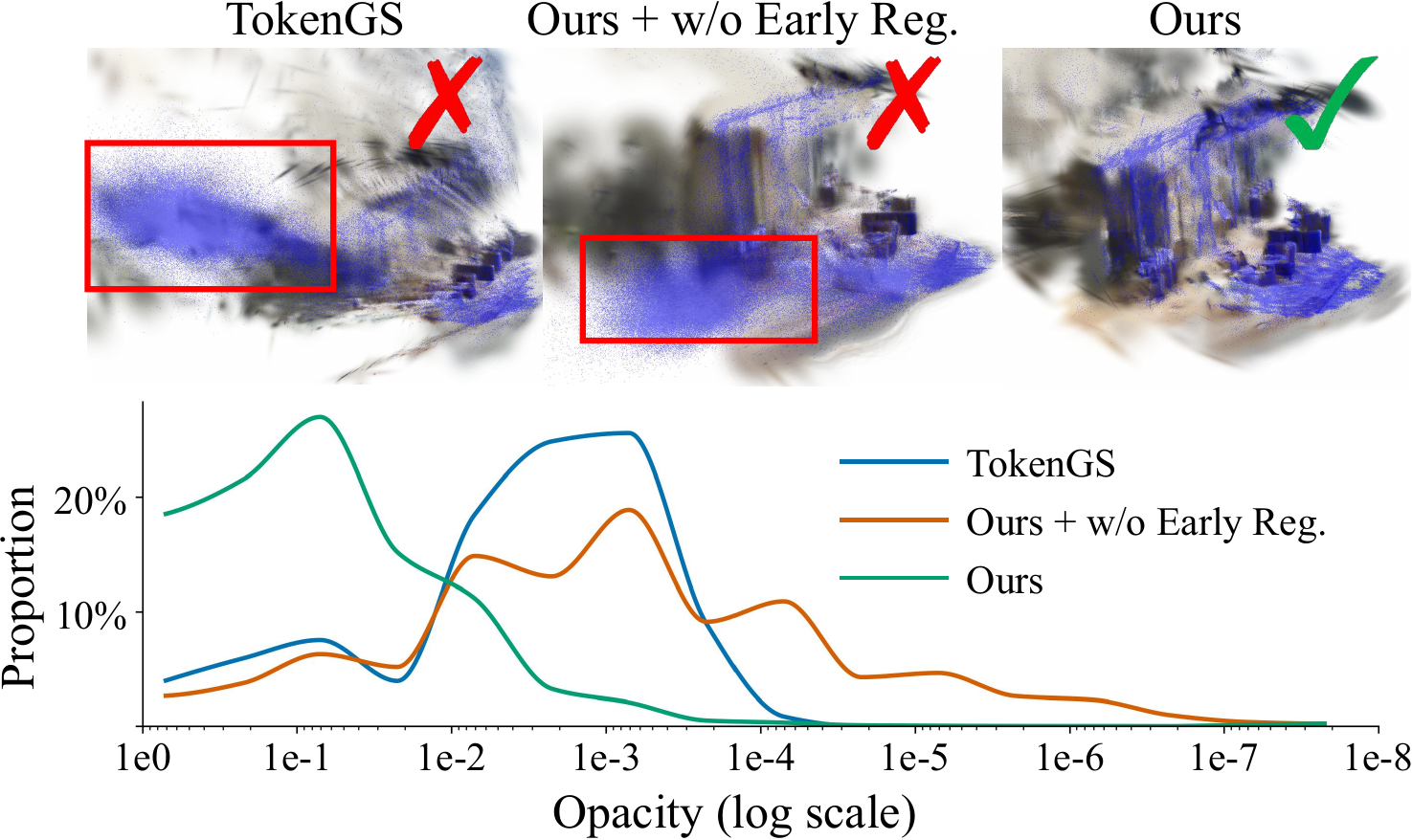}
    \caption{\textbf{Effect of early-stage regularization.} Compared with TokenGS and our variant without early-stage regularization, our model produces a cleaner Gaussian structure and a healthier opacity distribution. Red boxes highlight regions with concentrations of redundant and ineffective Gaussians.}
    \label{fig:alpha_distribution}
\end{figure}

\paragraph{Early-stage regularization.}
Table~\ref{tab:ablation-regularization} shows that early-stage regularization improves reconstruction quality by guiding unconstrained queries toward a plausible spatial and opacity configuration at the beginning of training. Consistently, Figure~\ref{fig:alpha_distribution} shows that the full model produces a cleaner Gaussian structure and a more balanced opacity profile than TokenGS~\cite{ren2026tokengs} and the variant without early-stage regularization. These results indicate that such transient guidance benefits both rendering quality and the geometric organization of the reconstructed Gaussian scene.

\begin{table}[!t]
  \centering
  {
  \fontsize{9pt}{10pt}\selectfont
  \setlength{\tabcolsep}{5pt}
  \begin{tabular}{l|ccc}
    \toprule
    Method & PSNR $\uparrow$ & SSIM $\uparrow$ & LPIPS $\downarrow$ \\
    \midrule
    \textbf{Ours} (VGGT-$\Omega$) & \underline{22.8963} & \underline{0.7381} & \textbf{0.2682} \\
    \textbf{Ours} (VGGT) & \textbf{23.0723} & \textbf{0.7388} & \underline{0.2749} \\
    \bottomrule
  \end{tabular}
  }
  \caption{VGM-backbone ablation at 4-view interpolation.}
  \label{tab:ablation-vgm-backbone}
\end{table}

\begin{figure}[!t]
    \centering
    \includegraphics[width=\linewidth]{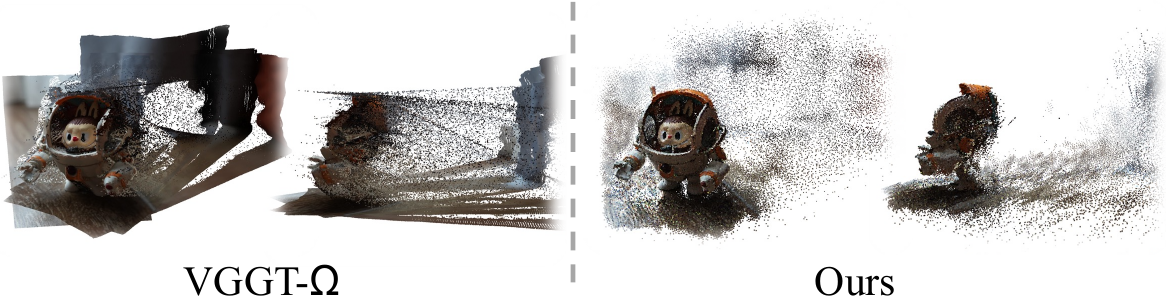}
    \caption{\textbf{Geometric comparison.}
    Left: point cloud obtained by back-projecting the depth predicted by VGGT-$\Omega$.
    Right: centers of the 3D Gaussians predicted by QuerySplat.}
    \label{fig:vgm_pointcloud_comparison}
\end{figure}

\paragraph{VGM backbone.}
Table~\ref{tab:ablation-vgm-backbone} examines whether the proposed decoder is tied to the particular geometric encoder used in the main model. Replacing the default backbone with VGGT~\cite{wang2025vggt} yields comparable quality without changing the decoder or training objective. This suggests that the effectiveness of QuerySplat primarily comes from how geometric priors are consumed, rather than from a model-specific feature representation. 

Beyond backbone compatibility, the learned query decoder also mitigates geometric artifacts inherited from the VGM. As shown in Figure~\ref{fig:vgm_pointcloud_comparison}, although the point cloud back-projected from VGGT-$\Omega$ depth contains noticeable floaters and noisy structures, the Gaussian centers predicted by QuerySplat exhibit a cleaner and more coherent spatial organization. This suggests that the decoder effectively reorganizes the VGM geometry prior under rendering supervision, rather than directly reproducing its raw predictions.

\subsection{Applications}
\label{sec:applications}

\paragraph{In-the-Wild Reconstruction.}
QuerySplat supports casually captured image sets with arbitrary view counts. As shown in Figure~\ref{fig:teaser}, it reconstructs clean and detailed 3D Gaussian scenes from one or multiple real-world images, enabling flexible everyday 3D capture.

\begin{figure}[!t]
    \centering
    \includegraphics[width=\linewidth]{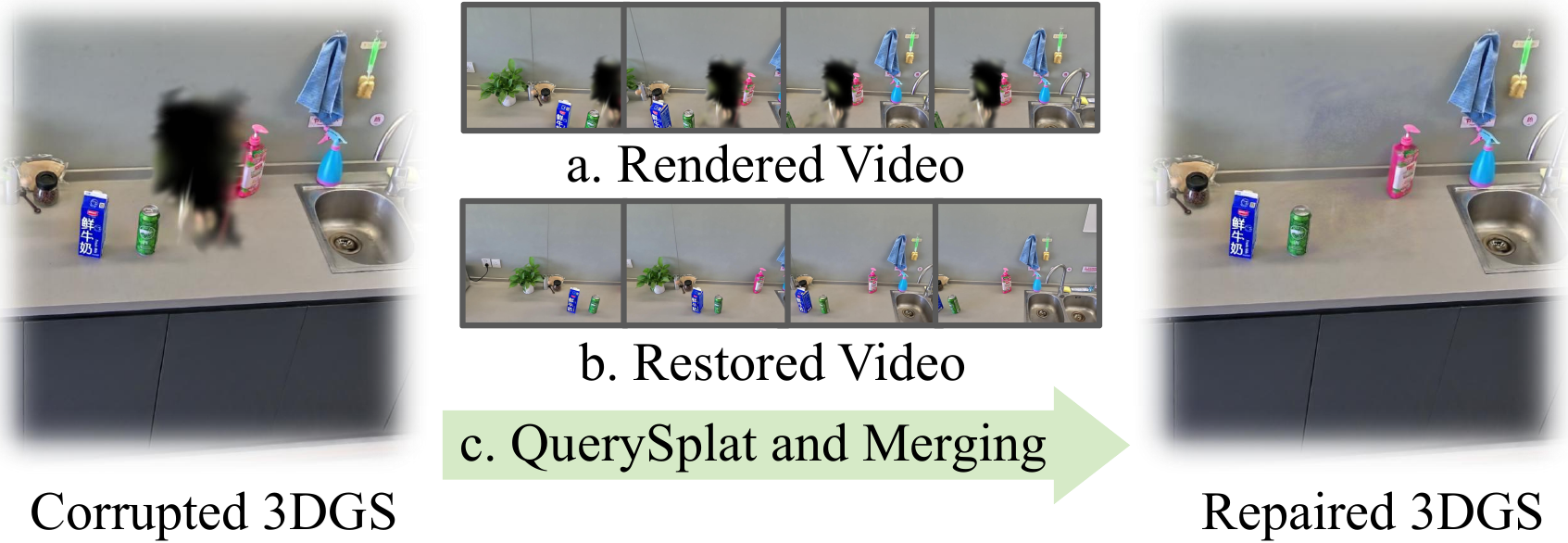}
    \caption{\textbf{3D scene repair.} Left: the manually corrupted Gaussian scene. Right: the reconstruction produced by QuerySplat after rendering the corrupted scene to video and applying video restoration.}
    \label{fig:scene_repair}
\end{figure}

\paragraph{3D Scene Repair.}
QuerySplat provides a rapid solution for 3D scene repair. We manually remove a region from an existing Gaussian scene and render the edited scene along a camera trajectory. Artifixer~\cite{de2026artifixer} repairs the missing content in the rendered video, after which QuerySplat reconstructs the restored frames into a new renderable Gaussian scene. As shown in Figure~\ref{fig:scene_repair}, the geometric prediction capability of QuerySplat enables the repaired 2D content to be lifted back into 3D with coherent structure and fewer floating artifacts, providing a practical way to integrate video restoration into 3D scene editing.

\begin{figure}[!t]
    \centering
    \includegraphics[width=0.9\linewidth]{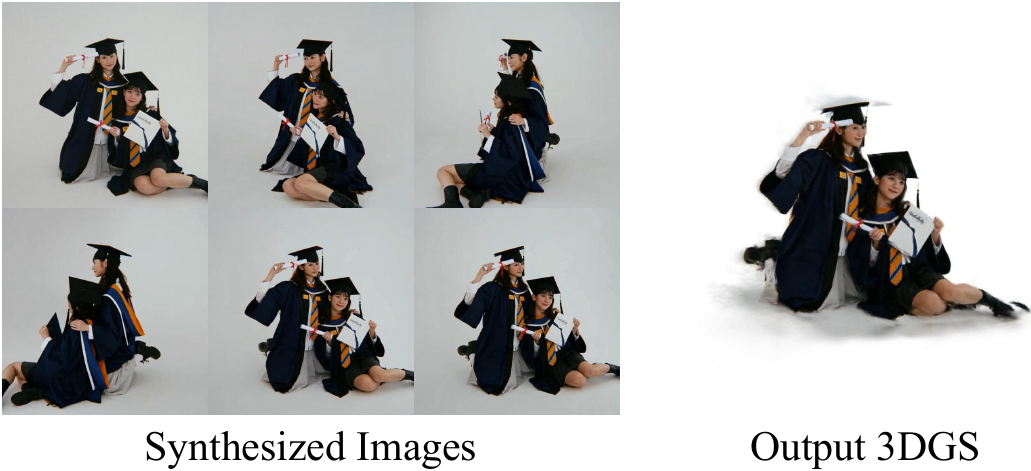}
    \caption{\textbf{3D reconstruction from T2V-generated views.} QuerySplat reconstructs a 3DGS scene from multi-view frames generated by a text-to-video model. The background is removed from the displayed 3DGS for visualization.}
    \label{fig:ai_generated}
\end{figure}

\paragraph{3D Reconstruction from T2V-Generated Views.}
We further evaluate QuerySplat on multi-view frames generated by a text-to-video model. These frames are directly used as unposed observations without any task-specific adaptation. As shown in Figure~\ref{fig:ai_generated}, QuerySplat reconstructs the generated visual content into a plausible and renderable 3D Gaussian scene. This result demonstrates that the proposed framework can operate not only on captured images, but also on synthetic multi-view content, providing a direct path from generative video outputs to explicit 3D scene representations.

\section{Conclusion}

We presented QuerySplat, a pose-free feed-forward 3DGS framework built around a dual-branch query decoder that separates geometric organization from high-frequency appearance modeling. The geometry branch establishes coherent scene structure, while the appearance branch restores fine visual details without sacrificing spatial flexibility. We instantiate the geometry branch with a VGM backbone to provide geometry-aware features and a self-consistent coordinate system, although the overall framework is not tied to a specific backbone. Progressive query scaling further increases representation capacity within the same reconstruction formulation. Extensive experiments show that QuerySplat produces cleaner Gaussian structures and sharper renderings, achieving state-of-the-art overall performance against both posed and pose-free baselines. Its compatibility with different geometric backbones and diverse applications further demonstrates the generality of the proposed framework.

\bibliography{aaai2027}


\def\AppendixIncluded{}
%
%
%

\ifdefined\AppendixIncluded
\else
  \documentclass[letterpaper]{article}
  \usepackage[submission]{aaai2027}

  \usepackage[hyphens]{url}
  \usepackage{graphicx}
  \urlstyle{rm}
  \def\UrlFont{\rm}
  \usepackage{natbib}
  \usepackage{caption}
  \frenchspacing
  \usepackage{booktabs}
   \usepackage{multirow} 

  \usepackage{amsmath}
  \usepackage{amssymb}

  \pdfinfo{
    /TemplateVersion (2027.1)
  }

  \setcounter{secnumdepth}{1}

  \title{Supplementary Material for\\QuerySplat}
  \author{Anonymous AAAI-27 Submission}
  \affiliations{Paper ID: XXXX}

  \begin{document}
  \maketitle
\fi

\ifdefined\AppendixIncluded
  \clearpage
\fi

\begin{center}
  {\LARGE\bfseries Appendix}
\end{center}

\appendix
\setcounter{secnumdepth}{1}

\section{Detailed Quantitative Comparisons}
\label{app:detailed_quantitative_comparisons}

\begin{table*}[!t]
  \centering
  \setlength{\tabcolsep}{1.5pt}
  \begin{tabular*}{\textwidth}{@{\extracolsep{\fill}}lc*{9}{c}@{}}
    \toprule
    \multirow{2}{*}{Method} & \multirow{2}{*}{Pose-free} & \multicolumn{3}{c}{input} & \multicolumn{3}{c}{interp} & \multicolumn{3}{c}{extrap} \\
    \cmidrule(lr){3-5}\cmidrule(lr){6-8}\cmidrule(lr){9-11}
    & & PSNR $\uparrow$ & SSIM $\uparrow$ & LPIPS $\downarrow$ & PSNR $\uparrow$ & SSIM $\uparrow$ & LPIPS $\downarrow$ & PSNR $\uparrow$ & SSIM $\uparrow$ & LPIPS $\downarrow$ \\
    \midrule
    DepthSplat~\shortcite{xu2025depthsplat} & $\times$ & 24.1130 & \underline{0.8729} & \textbf{0.1101} & 17.4135 & 0.5472 & \underline{0.3651} & 18.7962 & \underline{0.6150} & \textbf{0.3006} \\
    TokenGS~\shortcite{ren2026tokengs} & $\times$ & \underline{26.2841} & 0.8457 & 0.2626 & \underline{17.6985} & \underline{0.5703} & 0.4828 & \underline{18.9342} & 0.6123 & 0.4359 \\
    YoNoSplat~\shortcite{ye2025yonosplat} & $\times$ & 24.3481 & 0.8471 & 0.1291 & 16.8822 & 0.4994 & 0.4050 & 18.1934 & 0.5555 & 0.3476 \\
    \midrule
    AnySplat~\shortcite{jiang2025anysplat} & $\checkmark$ & \textbf{28.2681} & \textbf{0.9186} & \underline{0.1125} & 11.7463 & 0.3890 & 0.5317 & 12.7023 & 0.4229 & 0.4860 \\
    NoPoSplat~\shortcite{ye2025noposplat} & $\checkmark$ & 19.6465 & 0.6290 & 0.2816 & 16.3361 & 0.4798 & 0.4398 & 17.0561 & 0.5063 & 0.4073 \\
    SPFSplat~\shortcite{huang2025spfsplat} & $\checkmark$ & 22.4244 & 0.7106 & 0.1704 & 16.5557 & 0.4753 & 0.4200 & 17.4773 & 0.5173 & 0.3708 \\
    SplatWeaver~\shortcite{wan2026splatweaver} & $\checkmark$ & 24.2824 & 0.8356 & 0.1674 & 13.0674 & 0.4087 & 0.4561 & 13.8261 & 0.4368 & 0.4187 \\
    YoNoSplat~\shortcite{ye2025yonosplat} & $\checkmark$ & 18.6309 & 0.5367 & 0.3204 & 16.0399 & 0.4531 & 0.4317 & 16.6372 & 0.4727 & 0.4064 \\
    \textbf{Ours} & $\checkmark$ & 26.2565 & 0.8251 & 0.1581 & \textbf{18.6272} & \textbf{0.5945} & \textbf{0.3594} & \textbf{19.5802} & \textbf{0.6269} & \underline{0.3206} \\
    \midrule
    \textbf{Ours} + TTO20 & $\checkmark$ & 29.6810 & 0.8633 & 0.1091 & 18.9195 & 0.6124 & 0.3509 & 20.1278 & 0.6475 & 0.3036 \\
    \textbf{Ours} + TTO50 & $\checkmark$ & 30.2295 & 0.8681 & 0.0950 & 18.8219 & 0.6088 & 0.3497 & 20.0710 & 0.6449 & 0.3004 \\
    \bottomrule
  \end{tabular*}
  \caption{Per-scale results for \textbf{2-view} input on the \textbf{large} split. Higher PSNR/SSIM and lower LPIPS are better. \textbf{Bold} and \underline{underlined} values indicate the best and second-best results, respectively; TTO variants are excluded from this comparison.}
  \label{tab:appendix-2views-large}
\end{table*}

\begin{table*}[!t]
  \centering
  \setlength{\tabcolsep}{1.5pt}
  \begin{tabular*}{\textwidth}{@{\extracolsep{\fill}}lc*{9}{c}@{}}
    \toprule
    \multirow{2}{*}{Method} & \multirow{2}{*}{Pose-free} & \multicolumn{3}{c}{input} & \multicolumn{3}{c}{interp} & \multicolumn{3}{c}{extrap} \\
    \cmidrule(lr){3-5}\cmidrule(lr){6-8}\cmidrule(lr){9-11}
    & & PSNR $\uparrow$ & SSIM $\uparrow$ & LPIPS $\downarrow$ & PSNR $\uparrow$ & SSIM $\uparrow$ & LPIPS $\downarrow$ & PSNR $\uparrow$ & SSIM $\uparrow$ & LPIPS $\downarrow$ \\
    \midrule
    DepthSplat~\shortcite{xu2025depthsplat} & $\times$ & 25.5474 & \underline{0.8932} & \textbf{0.0940} & \underline{20.5908} & \underline{0.6802} & \underline{0.2509} & 20.4691 & \underline{0.6993} & \textbf{0.2360} \\
    TokenGS~\shortcite{ren2026tokengs} & $\times$ & 27.5959 & 0.8795 & 0.2222 & 20.3666 & 0.6761 & 0.3729 & \underline{20.5332} & 0.6871 & 0.3623 \\
    YoNoSplat~\shortcite{ye2025yonosplat} & $\times$ & 25.9272 & 0.8738 & 0.1101 & 20.0542 & 0.6412 & 0.2796 & 20.1932 & 0.6649 & 0.2657 \\
    \midrule
    AnySplat~\shortcite{jiang2025anysplat} & $\checkmark$ & \textbf{28.3469} & \textbf{0.9189} & \underline{0.1060} & 14.1915 & 0.4464 & 0.4452 & 14.0380 & 0.4595 & 0.4272 \\
    NoPoSplat~\shortcite{ye2025noposplat} & $\checkmark$ & 21.8738 & 0.7114 & 0.2020 & 19.2279 & 0.6047 & 0.3046 & 18.9084 & 0.6016 & 0.3103 \\
    SPFSplat~\shortcite{huang2025spfsplat} & $\checkmark$ & 23.2262 & 0.7494 & 0.1460 & 19.3380 & 0.5931 & 0.2944 & 18.6967 & 0.5818 & 0.3035 \\
    SplatWeaver~\shortcite{wan2026splatweaver} & $\checkmark$ & 24.6273 & 0.8375 & 0.1620 & 15.6790 & 0.4970 & 0.3644 & 15.2040 & 0.5005 & 0.3617 \\
    YoNoSplat~\shortcite{ye2025yonosplat} & $\checkmark$ & 21.9742 & 0.6774 & 0.1998 & 19.4209 & 0.6011 & 0.2952 & 19.0395 & 0.5880 & 0.3005 \\
    \textbf{Ours} & $\checkmark$ & \underline{28.2766} & 0.8856 & 0.1098 & \textbf{21.4607} & \textbf{0.7069} & \textbf{0.2501} & \textbf{21.2292} & \textbf{0.7026} & \underline{0.2526} \\
    \midrule
    \textbf{Ours} + TTO20 & $\checkmark$ & 32.4299 & 0.9184 & 0.0695 & 22.0407 & 0.7232 & 0.2398 & 21.9990 & 0.7264 & 0.2358 \\
    \textbf{Ours} + TTO50 & $\checkmark$ & 33.1205 & 0.9228 & 0.0599 & 21.9792 & 0.7215 & 0.2382 & 21.9669 & 0.7256 & 0.2336 \\
    \bottomrule
  \end{tabular*}
  \caption{Per-scale results for \textbf{2-view} input on the \textbf{medium} split. Higher PSNR/SSIM and lower LPIPS are better. \textbf{Bold} and \underline{underlined} values indicate the best and second-best results, respectively; TTO variants are excluded from this comparison.}
  \label{tab:appendix-2views-medium}
\end{table*}

\begin{table*}[!t]
  \centering
  \setlength{\tabcolsep}{1.5pt}
  \begin{tabular*}{\textwidth}{@{\extracolsep{\fill}}lc*{9}{c}@{}}
    \toprule
    \multirow{2}{*}{Method} & \multirow{2}{*}{Pose-free} & \multicolumn{3}{c}{input} & \multicolumn{3}{c}{interp} & \multicolumn{3}{c}{extrap} \\
    \cmidrule(lr){3-5}\cmidrule(lr){6-8}\cmidrule(lr){9-11}
    & & PSNR $\uparrow$ & SSIM $\uparrow$ & LPIPS $\downarrow$ & PSNR $\uparrow$ & SSIM $\uparrow$ & LPIPS $\downarrow$ & PSNR $\uparrow$ & SSIM $\uparrow$ & LPIPS $\downarrow$ \\
    \midrule
    DepthSplat~\shortcite{xu2025depthsplat} & $\times$ & 27.1695 & 0.9171 & \textbf{0.0759} & \underline{23.6028} & \underline{0.7890} & \textbf{0.1617} & 22.7682 & \textbf{0.7871} & \textbf{0.1683} \\
    TokenGS~\shortcite{ren2026tokengs} & $\times$ & \underline{28.5841} & 0.9012 & 0.1918 & 23.2123 & 0.7744 & 0.2792 & \underline{22.9806} & 0.7713 & 0.2849 \\
    YoNoSplat~\shortcite{ye2025yonosplat} & $\times$ & 27.1469 & 0.8997 & 0.0888 & 22.5705 & 0.7483 & 0.1882 & 22.2366 & 0.7554 & 0.1937 \\
    \midrule
    AnySplat~\shortcite{jiang2025anysplat} & $\checkmark$ & 28.5431 & \underline{0.9220} & 0.0958 & 16.2849 & 0.4986 & 0.3655 & 15.5915 & 0.4987 & 0.3663 \\
    NoPoSplat~\shortcite{ye2025noposplat} & $\checkmark$ & 23.8145 & 0.7917 & 0.1393 & 21.6310 & 0.7145 & 0.2047 & 20.9450 & 0.6944 & 0.2247 \\
    SPFSplat~\shortcite{huang2025spfsplat} & $\checkmark$ & 24.0475 & 0.7880 & 0.1183 & 21.4553 & 0.6937 & 0.1995 & 20.4444 & 0.6635 & 0.2264 \\
    SplatWeaver~\shortcite{wan2026splatweaver} & $\checkmark$ & 25.3171 & 0.8459 & 0.1521 & 18.3395 & 0.5917 & 0.2794 & 17.3749 & 0.5840 & 0.2929 \\
    YoNoSplat~\shortcite{ye2025yonosplat} & $\checkmark$ & 24.3293 & 0.7801 & 0.1322 & 22.0417 & 0.7184 & 0.1976 & 21.2538 & 0.6952 & 0.2143 \\
    \textbf{Ours} & $\checkmark$ & \textbf{29.7749} & \textbf{0.9276} & \underline{0.0774} & \textbf{24.0786} & \textbf{0.7956} & \underline{0.1660} & \textbf{23.3463} & \underline{0.7790} & \underline{0.1831} \\
    \midrule
    \textbf{Ours} + TTO20 & $\checkmark$ & 34.5038 & 0.9590 & 0.0409 & 25.0680 & 0.8148 & 0.1527 & 24.6639 & 0.8090 & 0.1630 \\
    \textbf{Ours} + TTO50 & $\checkmark$ & 35.2782 & 0.9627 & 0.0337 & 25.0299 & 0.8144 & 0.1512 & 24.6862 & 0.8096 & 0.1601 \\
    \bottomrule
  \end{tabular*}
  \caption{Per-scale results for \textbf{2-view} input on the \textbf{small} split. Higher PSNR/SSIM and lower LPIPS are better. \textbf{Bold} and \underline{underlined} values indicate the best and second-best results, respectively; TTO variants are excluded from this comparison.}
  \label{tab:appendix-2views-small}
\end{table*}

\begin{table*}[!t]
  \centering
  \setlength{\tabcolsep}{1.5pt}
  \begin{tabular*}{\textwidth}{@{\extracolsep{\fill}}lc*{9}{c}@{}}
    \toprule
    \multirow{2}{*}{Method} & \multirow{2}{*}{Pose-free} & \multicolumn{3}{c}{input} & \multicolumn{3}{c}{interp} & \multicolumn{3}{c}{extrap} \\
    \cmidrule(lr){3-5}\cmidrule(lr){6-8}\cmidrule(lr){9-11}
    & & PSNR $\uparrow$ & SSIM $\uparrow$ & LPIPS $\downarrow$ & PSNR $\uparrow$ & SSIM $\uparrow$ & LPIPS $\downarrow$ & PSNR $\uparrow$ & SSIM $\uparrow$ & LPIPS $\downarrow$ \\
    \midrule
    DepthSplat~\shortcite{xu2025depthsplat} & $\times$ & 23.9394 & \underline{0.8564} & \textbf{0.1284} & 20.2901 & \underline{0.6871} & \textbf{0.2503} & 18.2692 & \underline{0.6175} & \textbf{0.3123} \\
    TokenGS~\shortcite{ren2026tokengs} & $\times$ & 25.0249 & 0.8090 & 0.2876 & \underline{20.5588} & 0.6712 & 0.3830 & \underline{18.6832} & 0.6076 & 0.4375 \\
    YoNoSplat~\shortcite{ye2025yonosplat} & $\times$ & 23.9833 & 0.8206 & \underline{0.1561} & 20.3328 & 0.6550 & 0.2722 & 18.6229 & 0.5968 & 0.3329 \\
    \midrule
    AnySplat~\shortcite{jiang2025anysplat} & $\checkmark$ & \underline{25.2204} & \textbf{0.8636} & 0.1599 & 14.9064 & 0.4713 & 0.4283 & 13.2399 & 0.4421 & 0.4728 \\
    SplatWeaver~\shortcite{wan2026splatweaver} & $\checkmark$ & 23.0137 & 0.7963 & 0.1874 & 16.5145 & 0.5306 & 0.3495 & 14.3430 & 0.4712 & 0.4055 \\
    YoNoSplat~\shortcite{ye2025yonosplat} & $\checkmark$ & 21.0418 & 0.6457 & 0.2282 & 19.3526 & 0.5926 & 0.2979 & 17.7817 & 0.5316 & 0.3583 \\
    \textbf{Ours} & $\checkmark$ & \textbf{26.0338} & 0.8337 & 0.1699 & \textbf{21.8724} & \textbf{0.7169} & \underline{0.2593} & \textbf{19.6819} & \textbf{0.6432} & \underline{0.3277} \\
    \midrule
    \textbf{Ours} + TTO20 & $\checkmark$ & 30.2927 & 0.8931 & 0.1100 & 23.0259 & 0.7501 & 0.2325 & 20.5441 & 0.6794 & 0.3026 \\
    \textbf{Ours} + TTO50 & $\checkmark$ & 31.3715 & 0.9045 & 0.0907 & 23.0427 & 0.7505 & 0.2262 & 20.5315 & 0.6798 & 0.2966 \\
    \bottomrule
  \end{tabular*}
  \caption{Per-scale results for \textbf{4-view} input on the \textbf{large} split. Higher PSNR/SSIM and lower LPIPS are better. \textbf{Bold} and \underline{underlined} values indicate the best and second-best results, respectively; TTO variants are excluded from this comparison.}
  \label{tab:appendix-4views-large}
\end{table*}

\begin{table*}[!t]
  \centering
  \setlength{\tabcolsep}{1.5pt}
  \begin{tabular*}{\textwidth}{@{\extracolsep{\fill}}lc*{9}{c}@{}}
    \toprule
    \multirow{2}{*}{Method} & \multirow{2}{*}{Pose-free} & \multicolumn{3}{c}{input} & \multicolumn{3}{c}{interp} & \multicolumn{3}{c}{extrap} \\
    \cmidrule(lr){3-5}\cmidrule(lr){6-8}\cmidrule(lr){9-11}
    & & PSNR $\uparrow$ & SSIM $\uparrow$ & LPIPS $\downarrow$ & PSNR $\uparrow$ & SSIM $\uparrow$ & LPIPS $\downarrow$ & PSNR $\uparrow$ & SSIM $\uparrow$ & LPIPS $\downarrow$ \\
    \midrule
    DepthSplat~\shortcite{xu2025depthsplat} & $\times$ & 25.3594 & 0.8781 & \textbf{0.1082} & 22.7572 & \underline{0.7687} & \underline{0.1837} & 19.6293 & \underline{0.6801} & \textbf{0.2589} \\
    TokenGS~\shortcite{ren2026tokengs} & $\times$ & \underline{26.6703} & 0.8545 & 0.2355 & \underline{23.2286} & 0.7589 & 0.2981 & \underline{20.1766} & 0.6754 & 0.3753 \\
    YoNoSplat~\shortcite{ye2025yonosplat} & $\times$ & 25.8785 & 0.8683 & 0.1149 & 22.7899 & 0.7512 & 0.1944 & 19.8817 & 0.6614 & 0.2797 \\
    \midrule
    AnySplat~\shortcite{jiang2025anysplat} & $\checkmark$ & 26.2466 & \underline{0.8807} & 0.1343 & 17.4232 & 0.5484 & 0.3387 & 14.1144 & 0.4759 & 0.4210 \\
    SplatWeaver~\shortcite{wan2026splatweaver} & $\checkmark$ & 24.2414 & 0.8227 & 0.1632 & 19.1637 & 0.6300 & 0.2675 & 15.1613 & 0.5185 & 0.3555 \\
    YoNoSplat~\shortcite{ye2025yonosplat} & $\checkmark$ & 23.5423 & 0.7532 & 0.1578 & 21.9269 & 0.7041 & 0.2112 & 19.1994 & 0.6099 & 0.2977 \\
    \textbf{Ours} & $\checkmark$ & \textbf{28.0864} & \textbf{0.8957} & \underline{0.1146} & \textbf{24.6659} & \textbf{0.8084} & \textbf{0.1758} & \textbf{21.0778} & \textbf{0.7122} & \underline{0.2673} \\
    \midrule
    \textbf{Ours} + TTO20 & $\checkmark$ & 32.8387 & 0.9406 & 0.0675 & 26.2940 & 0.8379 & 0.1516 & 22.1078 & 0.7477 & 0.2431 \\
    \textbf{Ours} + TTO50 & $\checkmark$ & 34.0826 & 0.9500 & 0.0535 & 26.4224 & 0.8399 & 0.1462 & 22.1743 & 0.7500 & 0.2371 \\
    \bottomrule
  \end{tabular*}
  \caption{Per-scale results for \textbf{4-view} input on the \textbf{medium} split. Higher PSNR/SSIM and lower LPIPS are better. \textbf{Bold} and \underline{underlined} values indicate the best and second-best results, respectively; TTO variants are excluded from this comparison.}
  \label{tab:appendix-4views-medium}
\end{table*}

\begin{table*}[!t]
  \centering
  \setlength{\tabcolsep}{1.5pt}
  \begin{tabular*}{\textwidth}{@{\extracolsep{\fill}}lc*{9}{c}@{}}
    \toprule
    \multirow{2}{*}{Method} & \multirow{2}{*}{Pose-free} & \multicolumn{3}{c}{input} & \multicolumn{3}{c}{interp} & \multicolumn{3}{c}{extrap} \\
    \cmidrule(lr){3-5}\cmidrule(lr){6-8}\cmidrule(lr){9-11}
    & & PSNR $\uparrow$ & SSIM $\uparrow$ & LPIPS $\downarrow$ & PSNR $\uparrow$ & SSIM $\uparrow$ & LPIPS $\downarrow$ & PSNR $\uparrow$ & SSIM $\uparrow$ & LPIPS $\downarrow$ \\
    \midrule
    DepthSplat~\shortcite{xu2025depthsplat} & $\times$ & 27.2767 & \underline{0.9066} & \textbf{0.0846} & 25.6765 & \underline{0.8491} & \underline{0.1219} & 21.9588 & \underline{0.7686} & \textbf{0.1888} \\
    TokenGS~\shortcite{ren2026tokengs} & $\times$ & \underline{28.1302} & 0.8880 & 0.1959 & \underline{26.0239} & 0.8395 & 0.2267 & \underline{22.7399} & 0.7662 & 0.2930 \\
    YoNoSplat~\shortcite{ye2025yonosplat} & $\times$ & 27.3757 & 0.8971 & 0.0893 & 25.4238 & 0.8344 & 0.1297 & 21.9304 & 0.7491 & 0.2061 \\
    \midrule
    AnySplat~\shortcite{jiang2025anysplat} & $\checkmark$ & 27.6666 & 0.9005 & 0.1126 & 19.8517 & 0.6253 & 0.2583 & 15.9692 & 0.5350 & 0.3505 \\
    SplatWeaver~\shortcite{wan2026splatweaver} & $\checkmark$ & 25.8388 & 0.8562 & 0.1378 & 22.0457 & 0.7276 & 0.1975 & 17.5045 & 0.6112 & 0.2830 \\
    YoNoSplat~\shortcite{ye2025yonosplat} & $\checkmark$ & 25.6956 & 0.8274 & 0.1127 & 24.6138 & 0.7980 & 0.1410 & 21.3475 & 0.7114 & 0.2185 \\
    \textbf{Ours} & $\checkmark$ & \textbf{29.5741} & \textbf{0.9234} & \underline{0.0867} & \textbf{27.1912} & \textbf{0.8754} & \textbf{0.1179} & \textbf{23.2606} & \textbf{0.7886} & \underline{0.1942} \\
    \midrule
    \textbf{Ours} + TTO20 & $\checkmark$ & 34.5002 & 0.9602 & 0.0480 & 29.4373 & 0.9032 & 0.0933 & 24.7796 & 0.8243 & 0.1697 \\
    \textbf{Ours} + TTO50 & $\checkmark$ & 35.6475 & 0.9666 & 0.0384 & 29.6687 & 0.9059 & 0.0888 & 24.9192 & 0.8279 & 0.1643 \\
    \bottomrule
  \end{tabular*}
  \caption{Per-scale results for \textbf{4-view} input on the \textbf{small} split. Higher PSNR/SSIM and lower LPIPS are better. \textbf{Bold} and \underline{underlined} values indicate the best and second-best results, respectively; TTO variants are excluded from this comparison.}
  \label{tab:appendix-4views-small}
\end{table*}

\begin{table*}[!t]
  \centering
  \setlength{\tabcolsep}{1.5pt}
  \begin{tabular*}{\textwidth}{@{\extracolsep{\fill}}lc*{9}{c}@{}}
    \toprule
    \multirow{2}{*}{Method} & \multirow{2}{*}{Pose-free} & \multicolumn{3}{c}{input} & \multicolumn{3}{c}{interp} & \multicolumn{3}{c}{extrap} \\
    \cmidrule(lr){3-5}\cmidrule(lr){6-8}\cmidrule(lr){9-11}
    & & PSNR $\uparrow$ & SSIM $\uparrow$ & LPIPS $\downarrow$ & PSNR $\uparrow$ & SSIM $\uparrow$ & LPIPS $\downarrow$ & PSNR $\uparrow$ & SSIM $\uparrow$ & LPIPS $\downarrow$ \\
    \midrule
    DepthSplat~\shortcite{xu2025depthsplat} & $\times$ & 20.7272 & \underline{0.7537} & \textbf{0.2147} & 20.0607 & 0.6960 & \underline{0.2522} & 16.7422 & \underline{0.5426} & \textbf{0.3870} \\
    TokenGS~\shortcite{ren2026tokengs} & $\times$ & 20.7762 & 0.6591 & 0.4168 & 19.8281 & 0.6213 & 0.4378 & 16.8357 & 0.5252 & 0.5175 \\
    YoNoSplat~\shortcite{ye2025yonosplat} & $\times$ & 21.5871 & 0.7396 & \underline{0.2215} & \underline{20.9436} & \underline{0.6962} & \textbf{0.2474} & \underline{17.2244} & 0.5390 & \underline{0.3882} \\
    \midrule
    AnySplat~\shortcite{jiang2025anysplat} & $\checkmark$ & \underline{22.0788} & \textbf{0.7762} & 0.2323 & 17.9160 & 0.5682 & 0.3562 & 13.5825 & 0.4462 & 0.4965 \\
    SplatWeaver~\shortcite{wan2026splatweaver} & $\checkmark$ & 20.7351 & 0.7116 & 0.2398 & 18.7349 & 0.5964 & 0.3014 & 14.3457 & 0.4453 & 0.4361 \\
    YoNoSplat~\shortcite{ye2025yonosplat} & $\checkmark$ & 20.1787 & 0.6341 & 0.2581 & 19.9175 & 0.6219 & 0.2735 & 16.7819 & 0.4957 & 0.4039 \\
    \textbf{Ours} & $\checkmark$ & \textbf{22.7212} & 0.7347 & 0.2677 & \textbf{21.8494} & \textbf{0.7024} & 0.2895 & \textbf{18.2572} & \textbf{0.5938} & 0.3917 \\
    \midrule
    \textbf{Ours} + TTO20 & $\checkmark$ & 27.0716 & 0.8473 & 0.1826 & 24.6674 & 0.7913 & 0.2214 & 19.1717 & 0.6435 & 0.3579 \\
    \textbf{Ours} + TTO50 & $\checkmark$ & 28.5342 & 0.8769 & 0.1475 & 25.2006 & 0.8063 & 0.1984 & 19.2486 & 0.6479 & 0.3473 \\
    \bottomrule
  \end{tabular*}
  \caption{Per-scale results for \textbf{12-view} input on the \textbf{large} split. Higher PSNR/SSIM and lower LPIPS are better. \textbf{Bold} and \underline{underlined} values indicate the best and second-best results, respectively; TTO variants are excluded from this comparison.}
  \label{tab:appendix-12views-large}
\end{table*}

\begin{table*}[!t]
  \centering
  \setlength{\tabcolsep}{1.5pt}
  \begin{tabular*}{\textwidth}{@{\extracolsep{\fill}}lc*{9}{c}@{}}
    \toprule
    \multirow{2}{*}{Method} & \multirow{2}{*}{Pose-free} & \multicolumn{3}{c}{input} & \multicolumn{3}{c}{interp} & \multicolumn{3}{c}{extrap} \\
    \cmidrule(lr){3-5}\cmidrule(lr){6-8}\cmidrule(lr){9-11}
    & & PSNR $\uparrow$ & SSIM $\uparrow$ & LPIPS $\downarrow$ & PSNR $\uparrow$ & SSIM $\uparrow$ & LPIPS $\downarrow$ & PSNR $\uparrow$ & SSIM $\uparrow$ & LPIPS $\downarrow$ \\
    \midrule
    DepthSplat~\shortcite{xu2025depthsplat} & $\times$ & 21.8850 & \underline{0.7884} & \underline{0.1859} & 21.4410 & 0.7521 & \underline{0.2092} & 16.5801 & 0.5507 & \textbf{0.3843} \\
    TokenGS~\shortcite{ren2026tokengs} & $\times$ & 22.0150 & 0.7065 & 0.3710 & 21.4151 & 0.6862 & 0.3812 & 16.9313 & 0.5381 & 0.5099 \\
    YoNoSplat~\shortcite{ye2025yonosplat} & $\times$ & \underline{22.9007} & 0.7844 & \textbf{0.1858} & \underline{22.3512} & \underline{0.7550} & \textbf{0.2037} & \underline{17.4104} & \underline{0.5554} & \underline{0.3856} \\
    \midrule
    AnySplat~\shortcite{jiang2025anysplat} & $\checkmark$ & 22.7391 & \textbf{0.7900} & 0.2160 & 19.2000 & 0.6218 & 0.3124 & 13.1844 & 0.4489 & 0.4961 \\
    SplatWeaver~\shortcite{wan2026splatweaver} & $\checkmark$ & 21.4336 & 0.7356 & 0.2212 & 19.9685 & 0.6543 & 0.2628 & 13.8483 & 0.4492 & 0.4378 \\
    YoNoSplat~\shortcite{ye2025yonosplat} & $\checkmark$ & 21.4287 & 0.6928 & 0.2173 & 21.1921 & 0.6850 & 0.2277 & 16.9888 & 0.5155 & 0.3988 \\
    \textbf{Ours} & $\checkmark$ & \textbf{23.7809} & 0.7732 & 0.2312 & \textbf{23.2820} & \textbf{0.7589} & 0.2404 & \textbf{18.1704} & \textbf{0.5906} & 0.3979 \\
    \midrule
    \textbf{Ours} + TTO20 & $\checkmark$ & 28.2484 & 0.8739 & 0.1517 & 26.4901 & 0.8413 & 0.1741 & 19.1523 & 0.6443 & 0.3629 \\
    \textbf{Ours} + TTO50 & $\checkmark$ & 29.7052 & 0.8988 & 0.1214 & 27.1794 & 0.8562 & 0.1527 & 19.2620 & 0.6499 & 0.3517 \\
    \bottomrule
  \end{tabular*}
  \caption{Per-scale results for \textbf{12-view} input on the \textbf{medium} split. Higher PSNR/SSIM and lower LPIPS are better. \textbf{Bold} and \underline{underlined} values indicate the best and second-best results, respectively; TTO variants are excluded from this comparison.}
  \label{tab:appendix-12views-medium}
\end{table*}

\begin{table*}[!t]
  \centering
  \setlength{\tabcolsep}{1.5pt}
  \begin{tabular*}{\textwidth}{@{\extracolsep{\fill}}lc*{9}{c}@{}}
    \toprule
    \multirow{2}{*}{Method} & \multirow{2}{*}{Pose-free} & \multicolumn{3}{c}{input} & \multicolumn{3}{c}{interp} & \multicolumn{3}{c}{extrap} \\
    \cmidrule(lr){3-5}\cmidrule(lr){6-8}\cmidrule(lr){9-11}
    & & PSNR $\uparrow$ & SSIM $\uparrow$ & LPIPS $\downarrow$ & PSNR $\uparrow$ & SSIM $\uparrow$ & LPIPS $\downarrow$ & PSNR $\uparrow$ & SSIM $\uparrow$ & LPIPS $\downarrow$ \\
    \midrule
    DepthSplat~\shortcite{xu2025depthsplat} & $\times$ & 23.6793 & 0.8307 & \underline{0.1458} & 23.5338 & 0.8167 & \underline{0.1535} & 18.1796 & 0.6259 & \underline{0.3159} \\
    TokenGS~\shortcite{ren2026tokengs} & $\times$ & 24.4677 & 0.7886 & 0.2873 & 24.1851 & 0.7805 & 0.2898 & 18.8982 & 0.6230 & 0.4278 \\
    YoNoSplat~\shortcite{ye2025yonosplat} & $\times$ & \underline{25.2678} & \underline{0.8484} & \textbf{0.1275} & \underline{24.9030} & \underline{0.8313} & \textbf{0.1364} & \underline{19.1503} & \underline{0.6381} & \textbf{0.3110} \\
    \midrule
    AnySplat~\shortcite{jiang2025anysplat} & $\checkmark$ & 24.4494 & 0.8276 & 0.1723 & 21.7598 & 0.7102 & 0.2282 & 14.3189 & 0.5034 & 0.4269 \\
    SplatWeaver~\shortcite{wan2026splatweaver} & $\checkmark$ & 23.3236 & 0.7922 & 0.1740 & 22.5108 & 0.7541 & 0.1904 & 15.0812 & 0.5223 & 0.3682 \\
    YoNoSplat~\shortcite{ye2025yonosplat} & $\checkmark$ & 23.9189 & 0.7835 & 0.1489 & 23.7872 & 0.7791 & 0.1539 & 18.6901 & 0.6029 & 0.3227 \\
    \textbf{Ours} & $\checkmark$ & \textbf{26.1943} & \textbf{0.8488} & 0.1578 & \textbf{25.9700} & \textbf{0.8443} & 0.1591 & \textbf{19.9808} & \textbf{0.6705} & 0.3197 \\
    \midrule
    \textbf{Ours} + TTO20 & $\checkmark$ & 30.9625 & 0.9237 & 0.0949 & 29.8434 & 0.9090 & 0.1045 & 21.3198 & 0.7252 & 0.2827 \\
    \textbf{Ours} + TTO50 & $\checkmark$ & 32.4794 & 0.9413 & 0.0737 & 30.7796 & 0.9219 & 0.0882 & 21.5483 & 0.7338 & 0.2710 \\
    \bottomrule
  \end{tabular*}
  \caption{Per-scale results for \textbf{12-view} input on the \textbf{small} split. Higher PSNR/SSIM and lower LPIPS are better. \textbf{Bold} and \underline{underlined} values indicate the best and second-best results, respectively; TTO variants are excluded from this comparison.}
  \label{tab:appendix-12views-small}
\end{table*}

The main paper reports interpolation results averaged over the three sampling splits. Tables~\ref{tab:appendix-2views-large}, \ref{tab:appendix-2views-medium}, and \ref{tab:appendix-2views-small} present the complete 2-view results; Tables~\ref{tab:appendix-4views-large}, \ref{tab:appendix-4views-medium}, and \ref{tab:appendix-4views-small} report the corresponding 4-view results; and Tables~\ref{tab:appendix-12views-large}, \ref{tab:appendix-12views-medium}, and \ref{tab:appendix-12views-small} provide the 12-view evaluation. Each table includes input-view reconstruction, interpolation, and extrapolation. The large, medium, and small splits correspond to progressively narrower sampling intervals, allowing us to examine reconstruction under different degrees of viewpoint change. Taken together, these results reveal a consistent distinction between reproducing the observed images and recovering a scene representation that remains reliable beyond them.

\paragraph{Input and novel-view reconstruction.}
Pixel-aligned methods are naturally competitive on the input views because their Gaussian predictions retain a direct correspondence with observed pixels and camera rays. QuerySplat remains strong in this setting despite removing that constraint, showing that scene-level queries can preserve input fidelity without being anchored to individual observations. Its advantage becomes clearer on interpolation and extrapolation, where the rendering camera departs from the input views. Across different view counts and sampling intervals, QuerySplat exhibits substantially more stable novel-view performance than previous pose-free and query-based methods. This suggests that the predicted Gaussians form a coherent scene-level organization rather than merely reproducing appearance along the observed rays.

\paragraph{Effect of the number of input views.}
With only two views, reconstruction is highly under-constrained, yet QuerySplat can still establish a plausible geometric layout and recover useful appearance information. As the number of views increases, the model benefits from broader scene coverage and stronger cross-view constraints. The improvement from 2 to 4 views is particularly evident, while the 12-view results show that the query decoder can continue integrating denser observations without changing the overall formulation. QuerySplat also maintains its advantage across the three sampling splits, indicating that it generalizes to both nearby viewpoints and wider camera displacements.

\paragraph{Effect of test-time optimization.}
TTO adapts the extracted features using the available input images. Its largest gains appear on input-view reconstruction because supervision is applied directly at these views; importantly, the improvement also transfers to interpolation and extrapolation, indicating that TTO refines the underlying scene representation rather than simply memorizing the observations. The benefit generally grows with the number of input views, as additional images provide denser geometric constraints and more complete appearance coverage. Most improvements are already obtained with a small optimization budget, while further steps provide more gradual refinement. TTO is therefore an optional accuracy--runtime trade-off and is excluded from the main feed-forward comparison.

\paragraph{Limitations and future directions.}
The current model uses a fixed query budget, which bounds the number of Gaussians available to represent each scene. Consequently, very large or highly complex environments may be under-represented. This limitation concerns physical scene extent and complexity, rather than the large/medium/small evaluation splits. A promising extension is to divide a large environment into overlapping regions, reconstruct each region independently with QuerySplat, and then align and merge the resulting Gaussian sub-scenes, followed by optional global refinement.

\section{More Qualitative Comparisons}
\label{sec:more_qualitative}

\begin{figure*}[!t]
    \centering
    \includegraphics[width=\textwidth]{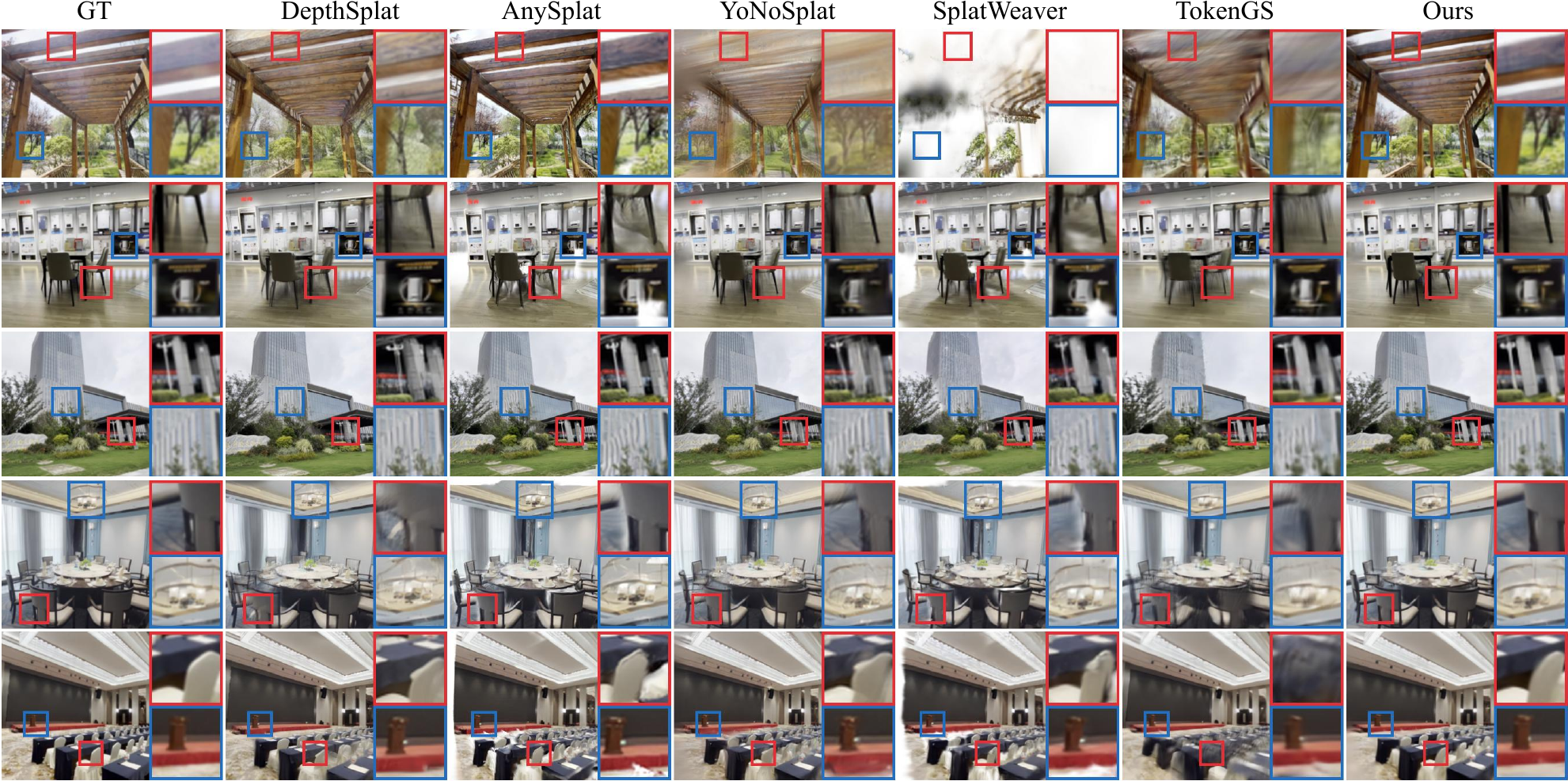}
    \caption{\textbf{Additional qualitative comparisons.}
    We compare QuerySplat with representative posed and pose-free feed-forward 3DGS methods on diverse scenes. The red and blue boxes highlight enlarged regions. QuerySplat preserves sharper boundaries, finer textures, and more coherent structures than competing methods.}
    \label{fig:more_qualitative_comparison}
\end{figure*}

\paragraph{Additional novel-view synthesis comparisons.}
Due to space limitations in the main paper, here, Figure~\ref{fig:more_qualitative_comparison} provides additional qualitative comparisons with representative feed-forward 3DGS methods. Across diverse indoor and outdoor scenes, QuerySplat preserves sharper object boundaries, thin structures, and high-frequency textures while reducing blur and structural distortions. The enlarged regions further demonstrate that our method recovers finer local details and produces renderings more consistent with the ground truth.

\begin{figure}[!t]
    \centering
    \includegraphics[width=\columnwidth]{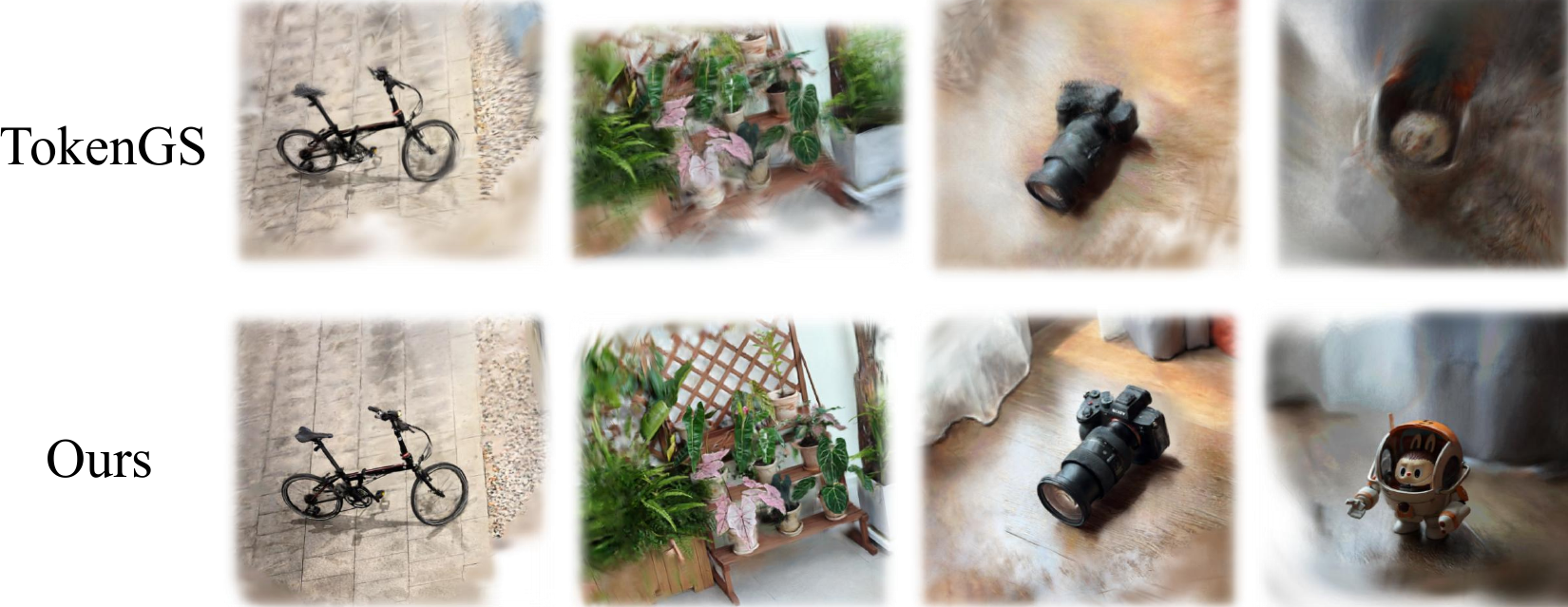}
    \caption{\textbf{Additional in-the-wild comparison with TokenGS.}
    TokenGS is a pose-required method and therefore receives cameras from an external VGGT-$\Omega$ predictor, followed by calibration to the DL3DV camera distribution used during its training. QuerySplat instead estimates cameras and establishes its coordinate system as part of its native pose-free reconstruction pipeline. Under this carefully calibrated adaptation, QuerySplat still preserves sharper object boundaries, finer appearance details, and more coherent structures. Because the TokenGS camera pipeline is not part of its original method, this result is presented as a supplementary comparison rather than a strictly like-for-like pose-free evaluation. (Note that TokenGS requires rectangular input, so the scene sizes are not strictly consistent)}
    \label{fig:in_the_wild_tokengs}
\end{figure}

\paragraph{In-the-wild comparison with a pose-required query model.}
In the main paper, we restrict the in-the-wild comparison to pose-free methods. This distinction is important because pose-required methods do not define how camera parameters should be obtained from uncalibrated images. Supplying externally estimated cameras introduces an additional camera-prediction and calibration system, making the final reconstruction dependent on components outside the original method. Consequently, comparisons between native pose-free methods and externally adapted pose-required methods are not strictly like-for-like. We therefore report the following comparison only as supplementary evidence rather than as part of the main in-the-wild evaluation.

To nevertheless compare against the most closely related query-based baseline, we adapt the released DL3DV 4-view checkpoint of TokenGS~\cite{ren2026tokengs} to uncalibrated images using an externally calibrated VGGT-$\Omega$~\cite{wang2026vggtomega} camera pipeline. Each original image is independently center-cropped and resized to $512\times512$ for VGGT-$\Omega$ camera prediction and to $256\times448$ for TokenGS inference, avoiding cascaded resizing between the two branches. VGGT-$\Omega$ predicts cameras under the OpenCV world-to-camera convention; we invert them to camera-to-world matrices and express all views relative to the first input camera, matching the relative camera convention used by TokenGS on DL3DV. To characterize the systematic discrepancy between the two camera distributions, we randomly sample 300 cases from DL3DV-Evaluation and process them using the same image preprocessing and view-sampling protocol. For each case, the predicted and ground-truth cameras are converted to the same convention and first-camera-relative frame, after which we compare their relative rotations and camera-center trajectories. The relative rotations are retained without additional correction, while a dataset-level translation-scale coefficient is robustly estimated from the ratio between the ground-truth and predicted trajectory spans across the sampled cases. During in-the-wild inference, the VGGT-$\Omega$ poses are first normalized to the reference camera, their relative rotations are kept unchanged, and their relative translations are rescaled by this precomputed coefficient to better match the DL3DV camera distribution seen during TokenGS training. We further transform the predicted intrinsics analytically from the VGGT-$\Omega$ crop coordinates to the TokenGS crop coordinates before constructing the Pl\"ucker ray embeddings. The resulting RGB images, calibrated poses, intrinsics, and rays are then passed to the released TokenGS model without finetuning.

As shown in Figure~\ref{fig:in_the_wild_tokengs}, QuerySplat produces substantially sharper and more coherent reconstructions across diverse casually captured scenes. TokenGS frequently loses high-frequency appearance and thin structures, resulting in blurred object boundaries and distorted local geometry. In contrast, QuerySplat more faithfully preserves the bicycle frame, individual plant leaves, the shape of the camera body, and the detailed appearance of the toy. These results further demonstrate the advantage of combining geometry-aware decoding with an explicitly separated appearance pathway.

\section{Inference Efficiency and Memory Usage}
\label{sec:inference_efficiency}

We benchmark the inference efficiency of QuerySplat on a single NVIDIA H200 GPU. Runtime is averaged over five CUDA-synchronized forward passes after one excluded warm-up pass. Peak GPU memory is measured for one complete 8,192-query forward pass using \texttt{torch.cuda.max\_memory\_allocated()} and reported in GiB. The VGGT-$\Omega$~\cite{wang2026vggtomega} time includes feature aggregation, while the decoder time excludes the encoder.

\begin{table*}[!t]
\centering
\begin{tabular}{cccccc}
\toprule
\shortstack{\# Views} &
\shortstack{VGGT-Omega\\Aggregator} &
\shortstack{Our Decoder\\(1024)} &
\shortstack{Our Decoder\\(8192)} &
\shortstack{Total Infer\\Time (8192)} &
\shortstack{Total (8192)\\Peak Memory (GiB)} \\
\midrule
1 & 0.080 & 0.071 & 0.621 & 0.703 & 9.276 \\
2 & 0.129 & 0.076 & 0.648 & 0.780 & 9.329 \\
3 & 0.205 & 0.080 & 0.675 & 0.884 & 9.383 \\
4 & 0.242 & 0.085 & 0.702 & 0.949 & 9.436 \\
5 & 0.345 & 0.090 & 0.729 & 1.082 & 9.489 \\
6 & 0.410 & 0.095 & 0.758 & 1.175 & 9.542 \\
8 & 0.550 & 0.105 & 0.818 & 1.377 & 9.648 \\
12 & 0.931 & 0.125 & 0.943 & 1.890 & 10.064 \\
24 & 2.559 & 0.182 & 1.293 & 3.880 & 11.549 \\
48 & 7.970 & 0.297 & 1.990 & 10.017 & 14.706 \\
72 & 16.393 & 0.411 & 2.687 & 19.158 & 18.431 \\
100 & 30.031 & 0.545 & 3.499 & 33.645 & 22.588 \\
200 & 111.950 & 1.025 & 6.400 & 118.574 & 29.979 \\
300 & 246.115 & 1.501 & 9.297 & 255.725 & 40.773 \\
\bottomrule
\end{tabular}
\caption{QuerySplat inference time (s) and 8192-query peak allocated GPU memory (GiB).}
\label{tab:inference_efficiency}
\end{table*}

As shown in Table~\ref{tab:inference_efficiency}, QuerySplat reconstructs scenes from up to four input views in under one second and processes 12 views in 1.89 seconds with the 8,192-query model. Peak memory remains around 10~GiB for up to 12 views and grows to 22.59 and 40.77~GiB for 100 and 300 views, respectively. Increasing the query count primarily affects decoder runtime, whereas the VGGT-$\Omega$ aggregation path becomes the dominant cost for large numbers of input views. These results show that QuerySplat supports both fast sparse-view reconstruction and substantially larger input collections on a single accelerator.

\section{Ablation Details}
\label{app:ablation_details}

\begin{figure*}[!t]
    \centering
    \includegraphics[width=\textwidth]{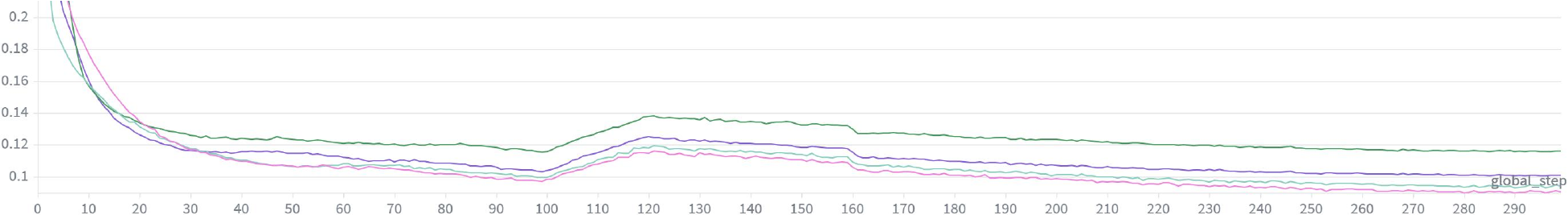}
    \caption{\textbf{Training-loss curves of the ablation variants.}
    In the later training stage, the curves correspond from top to bottom to \emph{w/o Appearance}, \emph{One-branch Query}, \emph{w/o Early Reg.}, and the full model. All variants exhibit similar convergence trends and maintain a stable relative ordering. The temporary increase in the middle is caused by the scheduled introduction of LPIPS, while the subsequent abrupt decrease results from a shared training-time filtering strategy. (Note that each global step contains 500 (0.5k) steps with totally 150k steps)}
    \label{fig:ablation_loss}
\end{figure*}

\paragraph{Training protocol.}
The main QuerySplat model is trained with a base stage followed by progressive finetuning with an increasing number of Gaussian queries. Fully repeating this training procedure for every ablation variant would incur prohibitive computational cost. Therefore, all ablation models are trained only in the base stage for 150K iterations, without progressive query expansion. Except for the component being ablated, all variants use the same training data, input-view sampling strategy, optimization settings. For the shortened 150K ablation runs, the scheduled loss transitions are correspondingly rescaled and kept identical across all variants. The quantitative ablations reported in the main paper are obtained from these base-stage checkpoints.

\paragraph{Convergence behavior.}
Figure~\ref{fig:ablation_loss} presents the training-loss curves of the four variants. In the later training stage, the curves correspond from top to bottom to \emph{w/o Appearance}, \emph{One-branch Query}, \emph{w/o Early Reg.}, and the full model. All variants exhibit closely aligned optimization dynamics: they decrease rapidly during early training, undergo the same scheduled transitions, and subsequently continue to converge at comparable rates. More importantly, the relative ordering between the variants becomes stable well before the end of base training, with no indication that the inferior variants are closing the performance gap. This suggests that extending all variants through the substantially more expensive progressive-finetuning stage would be unlikely to reverse the conclusions of the ablation study. The 150K-iteration base training therefore provides a sufficient and computationally practical comparison of the proposed components.

The temporary increase in loss observed in the middle of training is caused by the scheduled activation and gradual introduction of the LPIPS term. Because the plotted objective begins to include an additional perceptual loss component, its absolute value increases even though the underlying optimization remains stable. The later abrupt decrease is caused by a training-time filtering strategy shared by all variants. Since both changes occur consistently across the four models, they reflect common training-schedule transitions rather than instability introduced by any particular architectural design.

\section{Parameter Details}
\label{app:parameter_details}

\paragraph{Hardware and software environment.}
QuerySplat is trained on 64 NVIDIA A800 GPUs using distributed data parallelism. Each GPU processes one training sample, resulting in a global batch size of 64 without gradient accumulation. Training uses BF16 mixed precision, while the perceptual loss is evaluated in FP32 for improved numerical stability. Our implementation is based on Python 3.12 and PyTorch 2.11, with CUDA 12.8 and cuDNN 9.19, running on Ubuntu 24.04.

\paragraph{Model and training configuration.}
The input images are center-cropped and resized to $512\times512$. We use a frozen pretrained VGGT-$\Omega$ as the geometric encoder and fuse features extracted from its 4th, 11th, 17th, and 23rd intermediate layers. The geometry and appearance decoders contain 12 and 6 Transformer blocks, respectively. Both use an embedding dimension of 2,048, 16 attention heads, and an MLP expansion ratio of 4. The geometry decoder is initialized with 1,024 learnable queries, each of which predicts 64 Gaussian primitives. The predicted Gaussians use first-order spherical harmonics, and their activated scales are capped at $0.075$.

Base training uses four input views and is performed for 300K iterations. We then progressively double the number of queries until reaching 8,192, training each expansion stage for 30K iterations with randomly sampled 2--12 input views. We use AdamW with an initial learning rate of $10^{-4}$ during base training and $10^{-5}$ during progressive finetuning. A linear warm-up is followed by cosine learning-rate decay, and the global gradient norm is clipped to $1.0$. An exponential moving average of the model parameters is maintained with a decay factor of $0.9995$.

The reconstruction objective uses an $\ell_1$ photometric loss together with SSIM and LPIPS, whose weights are set to $0.2$ and $0.05$, respectively. LPIPS is introduced progressively after the initial photometric training stage. The visibility loss has a weight of $1.0$. The bidirectional Chamfer-distance loss and opacity-floor regularization are assigned initial weights of $1.0$ and $0.1$, respectively, and are annealed to zero during early training. The opacity floor is set to $0.1$. Cameras are normalized relative to the first input view, and the renderer uses near and far clipping planes of $0.025$ and $125.0$. During late training, a loss-rank filtering strategy retains $95\%$ of samples according to their historical reconstruction losses to reduce the influence of unstable training cases.

\ifdefined\AppendixIncluded
\else
  \bibliography{aaai2027}
  \end{document}
\fi

\end{document}